\pdfoutput=1

\documentclass[11pt]{article}
\usepackage[table]{xcolor} 

\usepackage[final]{acl}

\usepackage{times}
\usepackage{latexsym}

\usepackage[T1]{fontenc}

\usepackage[utf8]{inputenc}

\usepackage{microtype}

\usepackage{inconsolata}

\usepackage{graphicx}

\usepackage{array,multirow}
\usepackage{arydshln} 
\usepackage{amssymb}
\usepackage{pifont} 
\usepackage{float}
\usepackage{amsfonts} 
\usepackage{amsmath}
\DeclareMathOperator*{\argmax}{arg\,max}

\usepackage{booktabs} 
\usepackage{hyperref}
\usepackage{cleveref} 
\crefformat{section}{\S#2#1#3} 
\crefformat{subsection}{\S#2#1#3}
\crefformat{subsubsection}{\S#2#1#3}
\usepackage{url}
\usepackage{tikz-dependency}
\usepackage{mathabx}
\usepackage{tabularray}
\usepackage{tcolorbox}
\usepackage{lipsum}
\usepackage{etoolbox}
\AtBeginEnvironment{tcolorbox}{\small}
\makeatletter
\newcommand*\bigcdot{\mathpalette\bigcdot@{.5}}
\newcommand*\bigcdot@[2]{\mathbin{\vcenter{\hbox{\scalebox{#2}{$\m@th#1\bullet$}}}}}
\makeatother

\title{Delta-KNN: Improving Demonstration Selection in In-Context Learning for Alzheimer's Disease Detection}

\author{Chuyuan Li$^\delta$, Raymond Li$^\delta$, Thalia S. Field$^\gamma$, Giuseppe Carenini$^\delta$ \\
    $^\delta$ Department of Computer Science\\
    $^\gamma$ Vancouver Stroke Program and Division of Neurology, Faculty of Medicine\\
    The University of British Columbia\\
    V6T 1Z4, Vancouver, BC, Canada \\
    \texttt{\{chuyuan.li, thalia.field\}@ubc.ca, \{raymondl, carenini\}@cs.ubc.ca}
    }

\normalsize

\begin{document}
\maketitle

\begin{abstract}
Alzheimer’s Disease (AD) is a progressive neurodegenerative disorder that leads to dementia, and early intervention can greatly benefit from analyzing linguistic abnormalities. In this work, we explore the potential of Large Language Models (LLMs) as health assistants for AD diagnosis from patient-generated text using in-context learning (ICL), where tasks are defined through a few input-output examples. 
Empirical results reveal that conventional ICL methods, such as similarity-based selection, perform poorly for AD diagnosis, likely due to the inherent complexity of this task. To address this, we introduce Delta-KNN, a novel demonstration selection strategy that enhances ICL performance. Our method leverages a delta score to assess the relative gains of each training example, coupled with a KNN-based retriever that dynamically selects optimal ``representatives'' for a given input.  
Experiments on two AD detection datasets across three open-source LLMs demonstrate that Delta-KNN consistently outperforms existing ICL baselines. Notably, when using the Llama-3.1 model, our approach achieves new state-of-the-art results, surpassing even supervised classifiers.\footnote{Our code is available at \url{https://github.com/chuyuanli/Delta-KNN/}.}
\end{abstract}

\section{Introduction}
\label{sec:intro}

Large Language Models (LLMs), powered by advanced deep learning and vast cross-disciplinary training data, have transformed Natural Language Processing (NLP) \citep{zhao2023survey, fan2024bibliometric}. 
They show promise in specialized fields like clinical medicine and healthcare \citep{bubeck2023sparks, cui2024llms, belyaeva2023multimodal, jin2024health}. However, their ability to outperform traditional Artificial Intelligence (AI) in tasks requiring deep understanding and nuanced analysis remains uncertain \citep{wang2023can}.

\begin{figure}
    \centering
    \includegraphics[width=.9\columnwidth]{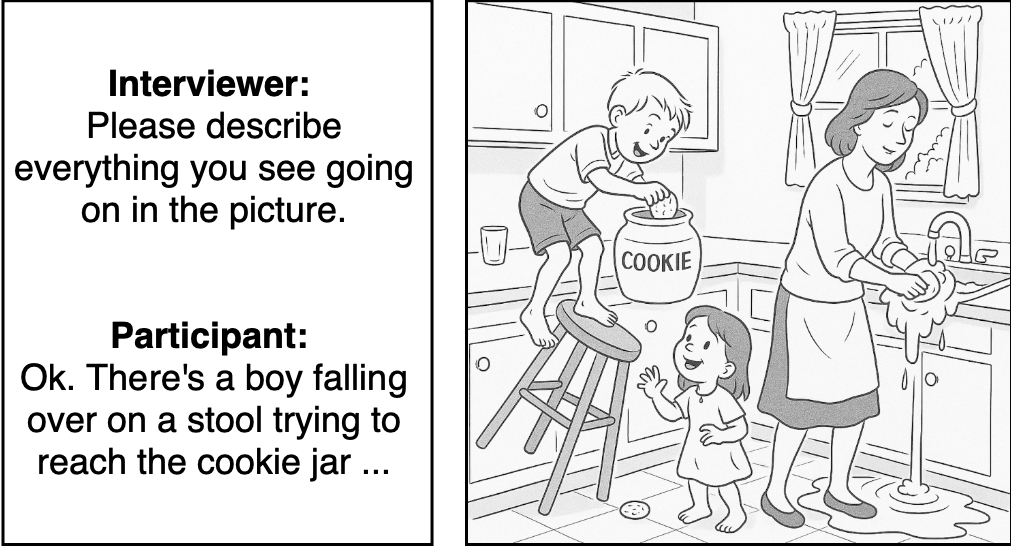}
    \caption[Picture generated using GPT-4o.]{The Cookie Theft picture\footnotemark description task.}
    \label{fig:cookie-theft}
    \vspace{-2ex}
\end{figure}
\footnotetext{This is an example picture created by \url{https://chatgpt.com}. The original picture, protected by \textcopyright\ 2000 PRO-ED, Incorporated, can be found in the BDAE-3 \citep{goodglass2001bdae}.
}

In this paper, we investigate LLM's capabilities in a crucial healthcare challenge: \textbf{Alzheimer’s Disease} detection. AD is a severe neurodegenerative disorder affecting 55 million people worldwide, ranking among the most costly diseases\footnote{\url{https://www.who.int/news-room/fact-sheets/detail/dementia}.}.  
Our approach focuses on identifying AD patients based on their descriptions of a given image, such as the well-known \textit{Cookie Theft} picture (Figure~\ref{fig:cookie-theft}). 
Traditional machine learning methods typically rely on feature extraction (e.g., linguistic analysis) \citep{fraser2016linguistic, fraser2019predicting, barral2020non} or embedding-based models (e.g., BERT) \citep{balagopalan2021comparing} to convert speech into vectors for classification. 
However, NLP has shifted from task-specific models to task-agnostic foundation models \citep{radford2019language, brown2020language}, enabling LLMs to not only simplify the diagnostic process but also offer interpretable reasoning, providing clinicians with valuable insights into their decision-making \citep{perlis2023application, nori2023capabilities, nori2023can}.

A major challenge in leveraging LLMs for AD detection is how to effectively teach them to learn from very \textbf{limited data} (i.e., few hundreds examples). 
In-context learning (ICL; \citep{brown2020language})---where a model performs a new task by conditioning on a few input-label pairs during inference---has emerged as a powerful and widely adopted strategy for handling complex tasks, which is applicable in data-poor scenarios.
One common approach involves \textit{similarity-based} selection \citep{liu2022whatmakes}, where examples resembling the target input or output are chosen. This method has shown strong performance in tasks like question answering, commonsense reasoning, and text-to-SQL generation \citep{liu2022whatmakes, su2023selective, raymond2025detriever}, but one concern is that the adopted similarity metrics may only capture a shallow understanding of the text. 
In order to enhance the model's \textit{understanding} of the target sample, \citet{peng2024revisiting} proposed a method that minimizes the conditional entropy between the demonstration and target input, demonstrating improvements on both classification and generation tasks. 

Other concerns include that ICL is highly sensitive to the selection of demonstration examples \citep{lu2022fantastically, iter2023context} and often struggles with tasks requiring complex reasoning \citep{peng2023does}. 
In light of these limitations, not surprisingly, our preliminary experiments reveal that existing ICL methods perform poorly on AD detection from text, which arguably requires the model to capture very subtle and complex linguistic and conceptual differences.

To address these challenges, we introduce a novel demonstration selection method, denoted as Delta-KNN, that practically quantifies the \textit{expected gain} of each example. This gain, referred to as the \textbf{delta score}, measures the improvement in model performance before and after including a demonstration. 
Using a small held-out set, we construct a \textbf{delta matrix} that stores performance gains for all examples. 
At inference time, we first identify target ``representatives'' by finding the \textbf{nearest neighbors} based on text similarity between the target sample and the held-out examples. 
Then, we select demonstrations that maximize the \textit{expected gain} for these representatives.
Extensive experiments on two AD detection datasets confirm the effectiveness of our approach, consistently outperforming existing demonstration selection methods. Additionally, we evaluate its robustness across different LLMs and explore its synergy with prompt engineering, achieving state-of-the-art (SOTA) performance comparable to supervised baselines.

In summary, (1) we introduce a \textbf{novel ICL method} designed to capture complex linguistic and conceptual nuances, making it particularly powerful in data-scarce scenarios;
(2) Our approach achieves \textbf{state-of-the-art performance}, surpassing existing ICL baselines in detecting dementia, which is one of the most costly diseases worldwide;
(3) Through extensive experiments, we show that the benefits of our method are conveniently \textbf{model- and prompt-agnostic}.

\section{Related Work}
\label{sec:related}

\paragraph{Language Analysis for AD detection. } 
Clinical studies have established a strong connection between speech and language abnormalities and AD pathology \citep{sajjadi2012abnormalities, rodriguez2016neuroanatomical}. 
Research in this area mostly relies on data from the Cookie Theft picture description task, particularly from the DementiaBank \citep{becker1994natural} and ADReSS \citep{luz2021detecting} datasets, and utilize semantic, syntactic, and lexical features \citep{ahmed2013connected, fraser2016linguistic, fraser2019predicting, jang2021classification}, with some studies also incorporating information unit analysis, such as counting object mentions in the picture \citep{masrani2017detecting, favaro2024}. 
While these methods achieve strong performance, they often rely on manual data annotation and feature engineering.
To reduce the need for labor-intensive processes, recent studies have explored deep learning approaches, including transfer learning \citep{zhu2021exploring, balagopalan2021comparing, agbavor2022predicting}, neural networks \citep{kong2019neural, fritsch2019automatic, bouazizi2023dementia}, and LLMs \citep{achiam2023gpt4, li2025largefoundation}. 
However, given the complexity of AD detection from text, naively prompting LLMs does not yield promising results \citep{wang2023can}. Instead, more sophisticated in-context learning strategies are required to fully explore LLM's inner specialist capabilities.

\begin{figure*}[]
    \centering
    \includegraphics[width=\textwidth]{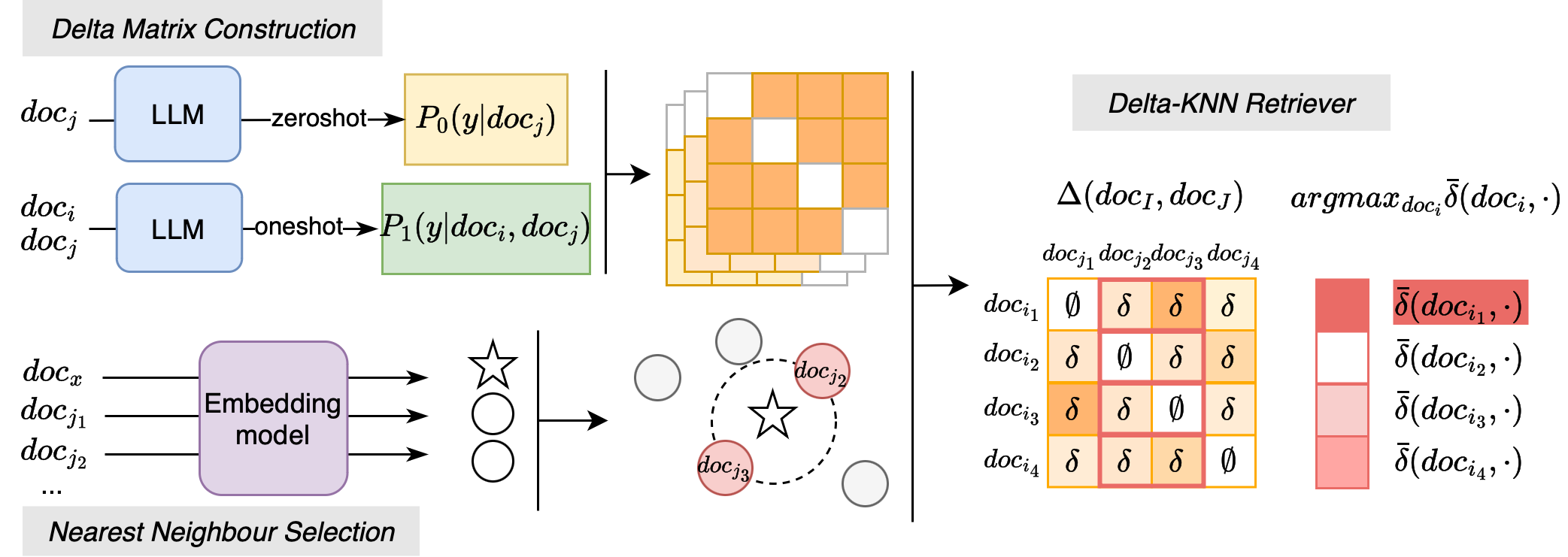}
    \caption{Delta-KNN retriever pipeline consists of two parts: (1) construct a delta matrix $\Delta$ by calculating the \textit{performance gain} from each demonstration example, where the gain is averaged over three runs to reduce model instability; (2) search for nearest neighbors (e.g., $doc_{j_2}, doc_{j_3}$) for target $doc_x$ in a vector space. The best demonstration example ($doc_{i_1}$) maximizes the average delta score over nearest neighbors ($\delta(doc_{i_1}, doc_{j_2}), \delta(doc_{i_1}, doc_{j_3})$).}
    \label{fig:pipeline}
\end{figure*}

\paragraph{Demonstration Selection in ICL.} 
Few-shot in-context learning (ICL) with LLMs has demonstrated performance comparable to supervised fine-tuning across various tasks like reasoning \citep{wei2022emergent, dong2022survey}. However, its effectiveness remains highly dependent on demonstration selection, leading to instability \citep{lu2022fantastically, peng2023does}. 
While \citet{lu2022fantastically} explored the impact of example order, they did not propose a method for selecting better examples. \citet{liu2022whatmakes} found that semantically similar examples improve ICL, later extended by incorporating more diverse demonstrations \citep{su2023selective}. 
Other studies have focused on enhancing model understanding through ranking mechanisms \citep{wu2023self}, perplexity-based prompt evaluation \citep{gonen2023demystifying}, and conditional entropy to assess model comprehension \citep{peng2024revisiting}. 
While these methods perform well on standard benchmarks, they remain untested in tasks like AD detection, where capturing subtle linguistic differences and reasoning-based cues is critical.

\section{Method}
\label{sec:method}

Our demonstration selection method consists of two modules: 
the first module constructs a \textit{performance gain} matrix using LLMs, referred to as the \textbf{Delta Matrix} ($\Delta$). Each cell in $\Delta$ contains a \textbf{delta score} ($\delta$) to represent the improvement or degradation contributed by a specific demonstration 
$doc_i$ to a target $doc_x$.
The second module employs an embedding-based retriever, which prioritizes demonstration examples based on their vector similarity to the target example. 
By integrating the outputs of both modules, we compute the average delta score ($\bar{\delta}$) for the $k$ nearest neighbors of each example in the training set. The optimal demonstration example is identified as the one with the highest aggregated delta score ($argmax(\bar{\delta})$). 
This process is illustrated in Figure~\ref{fig:pipeline}, and we describe each module in detail in the following sections.

\paragraph{Delta Matrix Construction with LLM.}
We construct the Delta Matrix by prompting the LLM in both zero-shot and one-shot scenarios. In the zero-shot scenario, the input to the LLM consists of the text from a document $doc_j$.
To elicit a probability alongside the predicted label, we include the cue phrase ``Give a prediction with a probability'' in the prompt, which has shown good calibration capabilities \citep{tian2023just}.
$P_0$ is the probability of the correct label for zero-shot prediction:
$\textstyle P_{0}(\hat{y} | doc_j; \theta)$ 
where $\theta$ refers to LLM parameters.

In the one-shot scenario, the predicted label and probability are obtained by appending the whole example $doc_i$ (text and label) prior to the text of $doc_j$. Similarly, we obtain the probability of correct prediction in one-shot: $\textstyle P_1(\hat{y} | doc_i, doc_j; \theta)$.

The \textbf{delta score} for a demonstration example $doc_i$ relative to the target $doc_j$ in the training set is defined as the difference between $P_1$ and $P_0$:
\begin{equation}\label{equ:1}
    \resizebox{\columnwidth}{!}{
    $\delta(doc_i,doc_j) = P_1(\hat{y} | doc_i, doc_j; \theta) - P_0(\hat{y} | doc_j; \theta).$}
\end{equation}

In a training set $D$ with $d$ number of documents,
the \textbf{Delta Matrix} $\Delta$ is a $d \times d$ matrix where each cell $[i, j]$ contains a delta score $\delta(doc_i, doc_j)$, representing the relative gain when $doc_i$ is used as a demonstration for $doc_j$: 
\begin{equation}\label{equ:5}
    \forall doc_i\in D, doc_j\in D, i \neq j, \Delta = \delta(doc_i,doc_j).
\end{equation}

\paragraph{Similarity-based $k$ Nearest Neighbors (KNN).}
We hypothesize that the average delta score $\bar{\delta}$ derived from guiding the documents \textbf{most similar} to the target document is more informative and effective.
Thus, we include a second module to select the most similar documents.
Specifically, we use a embedding model to convert documents to vector representations.
For each target $doc_x$ in the test set, we identify its $k$ nearest neighbors $doc_{j_1}$, $doc_{j_2}$, ..., $doc_{j_k}$ from the training set based on the distances in the embedding space. Using predefined similarity metrics, such as cosine similarity, the neighbors are ranked such that $doc_{j_1} < doc_{j_2}$ if $\cos(doc_{j_1}, doc_{x}) > cos(doc_{j_2}, doc_{x})$.
Note that the number of neighbors can vary: $k\in[1, d]$. In practice, we conduct simulations across a range of $k$ values within the training set using five-fold cross validation and apply the optimal $k$ value to the test documents.

A key element in this step is the \textbf{embedding model}, which projects sentences into a latent semantic space. 
We choose OpenAI embeddings \citep{openai2022embedding} as text encoder because it has shown superior results on a series of information retrieval tasks \citep{xian2024vector,lin2023aggretriever}, surpassing methods such as BM25 \citep{robertson2009probabilistic}, leaned sparse representations uniCOIL \citep{ma2022document}, and other semantic embedding APIs \citep{kamalloo2023evaluating}.
Precisely, we employ the latest \texttt{text-embedding-3-large} model\footnote{\url{https://openai.com/index/new-embedding-models-and-api-updates/}}.
Additionally, we compare this with representations obtained directly from LLMs (\cref{subsec:exp5}).

\paragraph{Delta-KNN Retriever.}
By combining the two modules described above, we compute the average delta score (Equation~\ref{equ:1}) of each demonstration $doc_i$ over the $k$ \textit{most similar} documents ($doc_{j_1}, ..., doc_{j_k}$) to $doc_{x}$: 
\begin{equation}\label{equ:6}
    \bar{\delta} (doc_{i}, \cdot) = \frac{1}{k} \sum_{k'=1}^{k} \delta(doc_{i}, doc_{j_{k'}}).
    \vspace{-1ex}
\end{equation}

This delta score represents the \textit{expected} gain when using $doc_{i}$ as a demonstration to the target $doc_{x}$.
Mathematically, we aim to find the best $doc_{i}$ by solving the following optimization problem: 
\begin{equation}\label{equ:7}
    doc_{i^*} = \argmax_{doc_{i}\in D} \bar{\delta}(doc_{i}, \cdot),
\end{equation}
where $doc_{i^*}$ is the example that maximizes the average delta score.
In $n$-shot ICL, we rank the examples in descending order and concatenate them to form the context $\{doc_{i_1}, ... doc_{i_n}\}$ prior to the target $doc_{x}$.

The Delta Matrix offers an intuitive map to guide the demonstration selection.
Different from existing similarity-based methods \citep{nori2023can} or text-understanding-based retrieval approaches \citep{peng2024revisiting}, our method is grounded in empirical evidence of performance gains observed from semantically similar documents.

\section{Experimental Setup}
\label{sec:setup}

\subsection{Datasets and Evaluation Metrics}
\label{subsec:datasets}

Picture Description Task, such as the one shown in Figure~\ref{fig:cookie-theft}, 
is a widely used task to capture deficits or abnormalities in language \citep{yorkston1980analysis, favaro2024}. 
In this work, we use two datasets that contain \textit{Cookie Theft} picture description for AD detection: ADReSS and Canary. 
\textbf{ADReSS} (Alzheimer’s Dementia Recognition through Spontaneous Speech) Challenge dataset \citep{luz2021detecting} is a curated subset of DementiaBank’s Pitt Corpus \citep{becker1994natural} that is matched for age and gender.
It consists of $156$ speech recordings and transcripts from AD (N~=~$78$) and non-AD (N~=~$78$) participants, and is divided into a training set and a test set.
\textbf{Canary} is collected by \citet{jang2021classification}, comprising $63$ patients recruited from a specialty memory clinic and $67$ healthy controls from the community. 
Patients are either diagnosed with AD or exhibiting initial symptoms of Mild Cognitive Impairments potentially progressing to AD. 
Canary includes longer documents with greater variations in length, gender, and age compared to ADReSS (see details in Appendix~\ref{append:data-process}), making it a more challenging dataset while more accurately reflecting clinically collected data.

For \textbf{evaluation metrics}, we use (1) Accuracy (ACC),
(2) Area Under the Curve (AUC) which captures the ability to distinguish between Patient and Control under different thresholds, (3) Sensitivity (SEN): the True Positive rate for Patient detection, and (4) Specificity (SPE): the True Negative rate for Control detection.

\begin{table*}[t!]
    \centering
    \resizebox{\textwidth}{!}{
    \begin{tabular}{lllllllllllll}
    \toprule
        \multirow{3}{*}{Method} & \multicolumn{4}{c}{ADReSS-train} & \multicolumn{4}{c}{ADReSS-test} &  \multicolumn{4}{c}{Canary}\\
        \cmidrule(lr){2-5} \cmidrule(lr){6-9} \cmidrule(lr){10-13}
        & ACC & AUC & SEN & SPE & ACC & AUC & SEN & SPE & ACC & AUC & SEN & SPE \\
        \midrule
    Zero-shot & $62.2\textsubscript{0.0}$ & $60.1\textsubscript{0.0}$ & $\mathbf{98.1\textsubscript{0.0}}$ & $22.2\textsubscript{0.0}$ & $57.6\textsubscript{1.0}$ & $57.6\textsubscript{1.0}$ & $\mathbf{100.0\textsubscript{0.0}}$ & $15.3\textsubscript{2.0}$ & $73.3\textsubscript{0.4}$ & $72.1\textsubscript{1.0}$ & $79.4\textsubscript{0.0}$ & $67.7\textsubscript{0.7}$ \\
    \hdashline
    Random & $68.4\textsubscript{2.2}$ & $71.9\textsubscript{3.1}$ & $84.0\textsubscript{2.3}$ & $48.8\textsubscript{6.3}$ & $75.7\textsubscript{4.3}$ & $81.5\textsubscript{2.6}$ & $93.1\textsubscript{2.0}$ & $58.3\textsubscript{9.0}$ & $73.1\textsubscript{2.7}$ & $75.3\textsubscript{3.7}$ & $72.0\textsubscript{3.3}$ & $74.1\textsubscript{2.5}$ \\
    Top-$k$ Select. & $69.0\textsubscript{1.6}$ & $71.9\textsubscript{2.5}$ & $88.3\textsubscript{2.3}$ & $45.7\textsubscript{1.7}$ & $70.1\textsubscript{2.0}$ & $80.0\textsubscript{0.8}$ & $91.7\textsubscript{3.4}$ & $48.6\textsubscript{2.0}$ & $71.0\textsubscript{2.5}$ & $75.0\textsubscript{2.2}$ & $76.7\textsubscript{0.7}$ & $65.7\textsubscript{4.2}$ \\
    ConE$^*$ Select. & $67.4\textsubscript{2.3}$ & $74.5\textsubscript{1.3}$ & $85.2\textsubscript{1.5}$ & $45.7\textsubscript{3.1}$ & $70.1\textsubscript{1.0}$ & $76.4\textsubscript{2.6}$ & $93.1\textsubscript{2.0}$ & $47.2\textsubscript{2.0}$ & $73.3\textsubscript{1.9}$ & $78.4\textsubscript{0.9}$ & $\mathbf{79.9\textsubscript{2.0}}$ & $67.2\textsubscript{4.4}$ \\
    \midrule
    Delta-KNN (ours) & $\mathbf{79.2\textsubscript{1.2}}$ & $\mathbf{78.9\textsubscript{1.3}}$ & $69.1\textsubscript{0.9}$ & $\mathbf{85.2\textsubscript{1.5}}$ & $\mathbf{80.5\textsubscript{3.9}}$ & $\mathbf{85.8\textsubscript{0.9}}$ & $70.8\textsubscript{5.9}$ & $\mathbf{86.1\textsubscript{2.0}}$ & $\mathbf{78.5\textsubscript{1.5}}$ & $\mathbf{79.8\textsubscript{0.9}}$ & $70.6\textsubscript{0.8}$ & $\mathbf{85.8\textsubscript{2.2}}$ \\
    \bottomrule
    \end{tabular}
    }
    \caption{AD detection results (accuracy, AUC, sensitivity, specificity) on ADReSS train set, ADReSS test set, and Canary using different demonstration selection methods. 
    We compare with zero-shot, random sampling, Top-$k$ \citep{liu2022whatmakes}, and ConE$^*$-based (conditional entropy) selection \citep{peng2024revisiting}.
    All results are averaged over three runs with standard deviation in subscription. Best score per column is in \textbf{bold}. 
    }
    \label{tab:exp1}
    \vspace{-1.5ex}
\end{table*}

\subsection{Baselines}
\label{subsec:baselines}
We compare our approach with popular demonstration selection methods. 
Since constructing the Delta Matrix relies on information from a training set, we also benchmark with supervised methods.

\paragraph{Demonstration Selection Methods.} Including:

\textbf{(1) Zero-Shot: }
A special case of ICL where no demonstration example is given.

\textbf{(2) Random Sampling: }
Randomly select examples for each target $i$.

\textbf{(3) Similarity-based Top-$k$ Selection: } Proposed in \citet{liu2022whatmakes} and has been widely used for health-related ICL \citep{nori2023can, nori2023capabilities}, where examples are embedded in a vector space and the nearest neighbors (calculated using cosine similarity) are selected as demonstration.

\textbf{(4) Text-understanding-based ConE Selection: }
A recent approach that quantifies \textit{understanding} by measuring the Conditional Entropy (ConE) of the target input given a demonstration and it selects examples that minimize the ConE \citep{peng2024revisiting}.

\paragraph{Supervised Baselines.} Including:

\textbf{(1) Statistical Machine Learning Classifiers: }
Traditional methods that use feature extraction (e.g., lexico-syntactic and semantic features) and supervised algorithms like Support Vector Space (SVM) \citep{luz2021detecting}, Random Forest (RF) \citep{luz2021detecting}, Logistic Regression (LR) \citep{jang2021classification}, and simple structure Neural Network (NN) \citep{balagopalan2021comparing}. 
We replicate a few studies and report results in \cref{subsec:exp7} and Appendix~\ref{append:superivsed}.

\textbf{(2) Transfer Learning-based Language Models: }
Pretrained Language Models (PLMs) like BERT \citep{devlin2018bert} encode rich linguistic information and are often fine-tuned for classification tasks without the need for manual feature extraction. 
We fine-tune a BERT model by following \citet{balagopalan2021comparing} (details in Appendix~\ref{append:ft-bert}) and include results from a SVM classifier which uses GPT-3 embeddings for contextualized input \citep{agbavor2022predicting}.

\textbf{(3) Supervised Fine-tuning: }
It is a common approach to adapt LLMs for downstream tasks by training on task-specific data, updating some or all parameters. 
We employ LoRA \citep{hulora2022}, a parameter-efficient fine-tuning strategy.

\textbf{(4) Learning-based Dense Retrievers: }
Using contrastive learning, the goal is to train a \textit{prompt retrieval} to score prompt based on some similarity metric.
We train a GPT-3 based language model following the Efficient Prompt Retrieval (EPR) method \citep{rubin2022learning}, as well as the Compositional Exemplars for In-context Learning (CEIL) approach \citep{ye2023compositional} where both diversity and similarity are considered for prompt selection.

\begin{table*}[t!]
    \centering
    \resizebox{\textwidth}{!}{
    \begin{tabular}{ccccccllllll}
        \toprule
        & \multirow{3}{*}{\texttt{Role}} & \multirow{3}{*}{\texttt{Con.}} & \multirow{3}{*}{\texttt{Ling.}} & \multirow{3}{*}{\texttt{CoT}} & \multirow{3}{*}{\texttt{G.-CoT}} & \multicolumn{2}{c}{ADReSS-train} & \multicolumn{2}{c}{ADReSS-test} & \multicolumn{2}{c}{Canary} \\
        \cmidrule(lr){7-12}
        && & & & & \multicolumn{2}{c}{$\Delta$-knn (Rdm, Top$k$, ConE)} & \multicolumn{2}{c}{$\Delta$-knn (Rdm, Top$k$, ConE)} & \multicolumn{2}{c}{$\Delta$-knn (Rdm, Top$k$, ConE)} \\
        \midrule
        (1) & \textcolor{red}{\ding{55}} & \textcolor{red}{\ding{55}} & \textcolor{red}{\ding{55}} & \textcolor{red}{\ding{55}} & \textcolor{red}{\ding{55}} 
        & $73.0$ &
        $\downarrow13.9$  
        $\downarrow17.6$
        $\downarrow12.7$
        & $69.8$ & $\uparrow0.3$ \hspace{2.72pt} $\downarrow2.8$ \hspace{2.72pt} $\uparrow0.8$
        & $63.1$ & $\downarrow2.3$ $\downarrow2.3$ $\downarrow3.9$ \\
        (2) & 
        \textcolor{blue}{\ding{51}} & \textcolor{blue}{\ding{51}} & \textcolor{red}{\ding{55}} & \textcolor{red}{\ding{55}} & \textcolor{red}{\ding{55}} 
        & $72.7$ & $\downarrow2.1$ \hspace{2.72pt} $\downarrow2.4$ \hspace{2.72pt} $\downarrow3.3$
        & $69.1$ & $\sim 0$ \hspace{8.72pt} $\sim 0$ \hspace{7.75pt} $\downarrow2.9$ 
        & $70.0$ & $\downarrow3.8$ $\downarrow3.8$ $\downarrow4.1$ \\
        (3) & 
        \textcolor{blue}{\ding{51}} & \textcolor{red}{\ding{55}} & \textcolor{blue}{\ding{51}} & \textcolor{red}{\ding{55}} & \textcolor{red}{\ding{55}} 
        & $73.1$ & $\downarrow7.8$ \hspace{2.72pt} $\downarrow13.1$  $\downarrow6.9$ 
        & $74.4$ & $\downarrow5.3$ \hspace{2.72pt} $\downarrow2.6$ \hspace{2.72pt} $\downarrow1.2$
        & $68.1$ & $\downarrow9.1$ $\downarrow2.2$ $\downarrow2.7$ \\
        (4) & 
        \textcolor{blue}{\ding{51}} & \textcolor{blue}{\ding{51}} & \textcolor{red}{\ding{55}} & \textcolor{blue}{\ding{51}} & \textcolor{red}{\ding{55}} 
        & $73.6$ & $\downarrow5.9$ \hspace{2.72pt} $\downarrow4.9$ \hspace{2.72pt} $\downarrow6.3$  
        & $74.6$ & $\downarrow2.1$ \hspace{2.72pt} $\downarrow2.8$ \hspace{2.72pt} $\downarrow2.8$ 
        & $71.5$ & $\downarrow9.7$ $\downarrow4.3$  $\downarrow10.2$ \\
        (5) & 
        \textcolor{blue}{\ding{51}} & \textcolor{red}{\ding{55}} & \textcolor{blue}{\ding{51}} & \textcolor{blue}{\ding{51}} & \textcolor{red}{\ding{55}} 
        & $74.5$ & $\downarrow10.2$ $\downarrow14.5$ $\downarrow16.3$
        & $74.6$ & $\downarrow11.1$  $\downarrow13.2$ $\downarrow16.7$ 
        & $65.1$ & $\downarrow3.6$ $\downarrow6.9$ $\downarrow7.7$ \\
        (6) & 
        \textcolor{blue}{\ding{51}} & \textcolor{blue}{\ding{51}} & \textcolor{blue}{\ding{51}} & \textcolor{blue}{\ding{51}} & \textcolor{red}{\ding{55}} 
        & $80.0$ & $\downarrow9.9$ \hspace{2.72pt} $\downarrow11.1$ $\downarrow8.6$ 
        & $83.6$ & $\downarrow13.8$  $\downarrow10.4$ $\downarrow12.5$ 
        & $70.8$ & $\downarrow7.5$ $\downarrow7.7$ $\downarrow9.6$ \\
        (7) & 
        \textcolor{blue}{\ding{51}} & \textcolor{blue}{\ding{51}} & \textcolor{blue}{\ding{51}} & \textcolor{red}{\ding{55}} & \textcolor{blue}{\ding{51}}
        & $79.2$ & $\downarrow10.8$ $\downarrow10.2$ $\downarrow11.8$ 
        & $80.5$ & $\downarrow2.8$ \hspace{2.72pt} $\downarrow8.4$ \hspace{2.72pt} $\downarrow8.4$ 
        & $78.5$ & $\downarrow5.4$ $\downarrow7.5$ $\downarrow5.2$ \\
        \bottomrule
    \end{tabular}
    }
    \caption{Delta-KNN ($\Delta$-knn) performance in accuracy on ADReSS and Canary using different prompting strategies: \texttt{Role}, \texttt{Context}, linguistic cues (\texttt{Ling}), chain-of-thought reasoning (\texttt{CoT}), and guided CoT (\texttt{G.-CoT}), in comparison with Random sampling (Rdm), Top-$k$ selection~\citep{liu2022whatmakes}, and Conditional Entropy (ConE)~\citep{peng2024revisiting} baselines. $\downarrow$, $\uparrow$, and $\sim$ symbols refer to lower, higher, and same accuracies compared to Delta-KNN, respectively.}
    \label{tab:exp2}
\end{table*}

\subsection{Implementation Details}
\label{subsec:implement}

Our experiments are conducted on Llama-3.1-8B-Instruct \citep{dubey2024llama}.
To assess the robustness of our method, we also test it on Qwen2.5-7B-Instruct \citep{yang2024qwen2} and Mistral-7B-Instruct-v0.3 \citep{jiang2023mistral}, see \cref{subsec:exp6}.

For zero-shot and few-shot ICL, we use a low temperature ($0.01$) and set top\_k sampling to $50$.
We use $4$-shot learning with two positive and two negative examples. The impact of in-context examples is discussed in \cref{subsec:exp4}.
To address the potential \textit{non-determinism} and instability of LLMs \citep{ouyang2023llm,song2024good}, each experiment is repeated three times, and we present the average scores with the standard deviation where applicable.
The Delta Matrix is constructed using the averaged zero-shot and one-shot results, as illustrated by the three matrices shown in Figure~\ref{fig:pipeline}.
For LLM fine-tuning, we employ LoRA technique \citep{hulora2022} and train the model for one epoch, with details in Appendix~\ref{append:ft-llama}.

Given the complexity of this task \citep{bouazizi2023dementia, favaro2024}, we carefully design prompts with comprehensive instructions to enhance the model's understanding and diagnostic capabilities on AD, following \citet{li2025largefoundation}. 
Our prompt includes: \texttt{Role}---``\textit{You are a medical expert in Alzheimer's Disease}'' to establish domain expertise, 
\texttt{Context}---a concise introduction to the Cookie Theft picture description task, and \texttt{Linguistic}---key linguistic features the model should focus on.  
In addition, we incorporate a guided Chain of Thought (\texttt{G.-CoT}) reasoning step \citep{kojima2022large}, prompting the model to analyze specific linguistic aspects such as \textit{vocabulary richness} and \textit{syntactic complexity}, supported by clinical observations \citep{ash2015study, forbes2005detecting, bouazizi2023dementia}. 
A complete few-shot prompt template is presented as: (\texttt{Role+Context+Linguistic}; \texttt{Demonstrations}; \texttt{G.-CoT}), and is provided in Appendix~\ref{append:prompts}.  
Preliminary zero-shot experiments validate the effectiveness of this prompt design \citep{li2025largefoundation}. To further analyze its impact, we introduce variations by ablating different components and evaluating them in \cref{subsec:exp2}.

\section{Experiments}
\label{sec:experiments}

We conduct experiments to show the effectiveness of Delta-KNN compared with other ICL methods (\cref{subsec:exp1}) and examine the influence of prompt engineering (\cref{subsec:exp2}), demonstration ordering (\cref{subsec:exp8}), and hyperparameters (\cref{subsec:exp4}, \cref{subsec:exp3}).

\subsection{Delta-KNN \textit{vs.} Other Demonstration Selection Methods}
\label{subsec:exp1}

Table~\ref{tab:exp1} presents AD detection results on the ADReSS and Canary datasets using Random sampling, Top-$k$, and ConE-based selection. 
We also report zero-shot results.
We prompt LLM with the (\texttt{Role+Context+Linguistic}; \texttt{Demonstrations}; \texttt{G.-CoT}) template, with all ICL methods containing four demonstrations with balanced labels (\cref{subsec:exp4}). The $k$ value in Delta-KNN is set to $13$, which is based on empirical results on training sets (\cref{subsec:exp3}).

A first interesting observation from Table~\ref{tab:exp1} is that zero-shot prompting almost always predicts participants as Patients, achieving as high as $100\%$ sensitivity while failing to identify Controls, a \textbf{bias} particularly evident in the ADReSS dataset.  
With in-context learning, models exhibit more balanced predictions---most ICL methods significantly improve specificity, with higher accuracy and AUC scores. This shows that learning from examples helps correct the model’s initial bias. 
Random sampling performs well overall, suggesting that exposure to a diverse input distribution benefits ICL \citep{nori2023capabilities}.
Surprisingly, the recent ConE-based selection delivers mixed results. While it improves performance on Canary, it falls short on ADReSS compared to Random sampling and Top-$k$ selection.  
In contrast, our proposed method consistently outperforms all selection methods on both datasets, achieving a $5-10\%$ and $5\%$ accuracy improvement on ADReSS and Canary, respectively. 
Notably, Delta-KNN excels at identifying speech from healthy controls (SPE: $70-85\%$) while maintaining strong performance in detecting patients (SEN: $70\%$). 
Overall, our method attains an optimal AUC score ($79-85\%$), highlighting the strong discriminative power of the selected examples.

\begin{table*}[t!]
    \centering
    \small
    \resizebox{\textwidth}{!}{
    \begin{tabular}{lcccccccccccc}
    \toprule
        \multirow{3}{*}{Method} & \multicolumn{4}{c}{ADReSS-train} & \multicolumn{4}{c}{ADReSS-test} &  \multicolumn{4}{c}{Canary}\\
        \cmidrule(lr){2-5} \cmidrule(lr){6-9} \cmidrule(lr){10-13}
        & Min & Mean & Std & Max & Min & Mean & Std & Max & Min & Mean & Std & Max \\
        \midrule
    Random & $63.9$ & $68.3$ & $1.9$ & $72.2$ & $68.8$ & $75.0$ & $4.1$ & $81.2$ & $66.9$ & $71.0$ & $2.2$ & $75.4$ \\
    Top-$k$ Select. & $60.2$ & $65.6$ & $2.6$ & $69.4$ & $66.7$ & $71.9$ & $2.8$ & $77.1$ & $65.4$ & $70.4$ & $2.0$ & $73.1$ \\
    ConE$^*$ Select. & $62.0$ & $64.9$ & $1.9$ & $67.6$ & $64.6$ & $71.1$ & $2.6$ & $75.0$ & $69.2$ & $72.3$ & $1.6$ & $74.6$ \\
    \midrule
    Delta-KNN (ours) & $\mathbf{73.2}$ & $\mathbf{79.2}$ & $\mathbf{1.8}$ & $\mathbf{80.6}$ & $\mathbf{76.0}$ & $\mathbf{80.9}$ & $\mathbf{2.5}$ & $\mathbf{83.3}$ & $\mathbf{75.3}$ & $\mathbf{78.1}$ & $\mathbf{0.9}$ & $\mathbf{79.2}$ \\
    \bottomrule
    \end{tabular}
    }
    \caption{
    The minimum (min), average (mean) with standard deviation (std), maximum (max) accuracies across all $24$ possible orderings in the 4-shot ICL setting, comparing our method (using full prompt: \texttt{Role+Context+Linguistic}; \texttt{Demonstrations}; \texttt{G.-CoT}) against random sampling, Top-$k$ \citep{liu2022whatmakes}, and ConE$^*$-based \citep{peng2024revisiting} methods. 
    Best score per column is in \textbf{bold}.
    }
    \label{tab:exp8}
\end{table*}

\subsection{Impact of Prompting Engineering}
\label{subsec:exp2}

It is known that a model's performance can be significantly affected by its prompt \citep{feng2024don, sivarajkumar2024empirical, sclarquantifying}. 
To examine the impact of prompt engineering and assess the robustness of our approach, we conduct ablation studies on prompt engineering. 
Precisely, we systematically vary the prompt design by gradually removing task-related information, ranging from a minimal prompt (lacking background details and CoT reasoning cues) to a comprehensive prompt containing all key components, i.e., (\texttt{Role+Context+Linguistic}; \texttt{G.-CoT}). 
 
In Table~\ref{tab:exp2}, we test seven variations using Delta-KNN, comparing it with Random sampling, Top~$k$, and ConE-based selection. The results clearly show that task-related information is crucial:  
without prompt engineering (Prompt 1), Delta-KNN achieves $69\%$ accuracy on ADReSS-test and $63\%$ on Canary, which are $11$ and $15$ points lower than the best-performing design (Prompt 7).  
Adding background details such as \texttt{Role}, \texttt{Context}, and \texttt{Linguistic} (Prompts 2 and 3) improves accuracy by $5\%$, confirming the importance of domain-specific context.  
When including a simple Chain-of-Thought (\texttt{CoT}) cue phrase ``\textit{First explain step-by-step and then give a prediction.}'', prompts 4 and 5 give further gains. Although marginal, it significantly enhances interpretability by making the model’s reasoning more transparent. 
Finally, combining all background information with \texttt{CoT} (Prompt 6) boosts performance, with the highest accuracy achieved using our \texttt{Guided CoT} (Prompt 7).  
Remarkably, across all prompt settings, Delta-KNN consistently outperforms other demonstration selection methods, demonstrating its robustness under different prompting strategies.

\begin{figure}[b]
    \centering
    \includegraphics[width=\columnwidth]{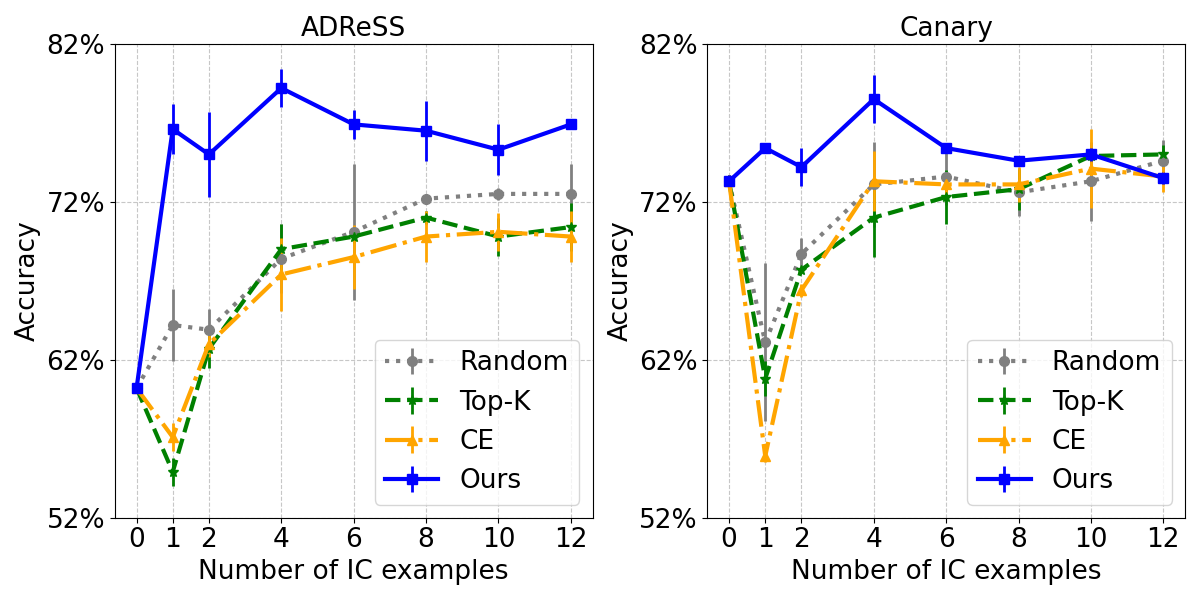}
    \caption{Impact of the number of in-context examples on ADReSS (left) and Canary (right) train sets. 
    }
    \label{fig:exp4}
\end{figure}

\subsection{Impact of Demonstration Ordering}
\label{subsec:exp8}

Previous research has demonstrated that in-context learning (ICL) is highly sensitive to the ordering of in-context examples \citep{lu2022fantastically, luo2024context}. 
To assess the sensitivity of our approach to this factor, we conduct an experiment evaluating its performance across all $24$ possible orderings in the 4-shot ICL setting.
Our method employs the Llama-3.1-8B-Instruct model along with the optimal prompting strategy described in \cref{subsec:exp2}, and is compared against random sampling, Top-$k$, and conditional entropy baselines.

As shown in Table~\ref{tab:exp8}, our method consistently achieves higher maximum and average accuracy scores than all baselines, with also lower standard deviation. 
These results suggest that our approach effectively reduces the impact of order sensitivity when using multiple demonstrations.

\subsection{Impact of In-Context Examples}
\label{subsec:exp4}

To assess the impact of the number of demonstrations, we gradually increase the number of examples ($N$) from 0 to 12.
As shown in Figure~\ref{fig:exp4}, performance generally improves with more examples.  

Interestingly, when using only one example, most selection methods experience a sharp performance drop compared to zero-shot, likely due to biased label distribution in demonstrations \citep{min2022rethinking}. In contrast, Delta-KNN outperforms zero-shot, indicating its ability to select the most beneficial example (i.e., with the highest delta score) for the target input. 
When demonstrations include a balanced mix of positive and negative samples, Top-$k$, Random Sampling, and ConE-based selection show improvements, particularly on ADReSS. However, on Canary, few-shot only begins win over zero-shot when $N \geq 4$.  
Across datasets, in-context performance increases, peaking at $N=4$, after which it fluctuates and stabilizes. 
Thus, we select four in-context examples for our experiments.

Although modern LLMs like Llama support long context windows, increasing the number of examples does not necessarily improve performance, indicating difficulty in focusing on the most informative content.
In our preliminary experiments, baseline methods with larger numbers of demonstrations ($N=20, 50, 80$) fall short compared to our $4$-shot Delta-KNN approach, with detailed analysis provided in Appendix~\ref{append:80examples}.

\subsection{Impact of $k$ value in Delta-KNN}
\label{subsec:exp3}

\begin{figure}[t]
    \centering
    \includegraphics[width=.85\columnwidth]{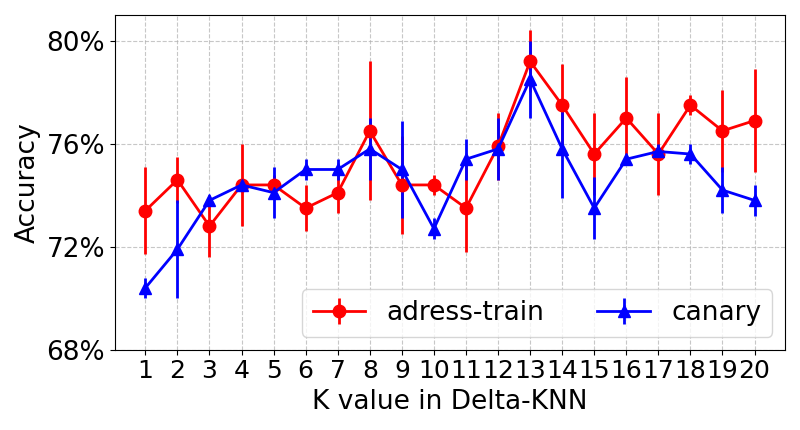}
   \caption{Impact of $k$ value in Delta-KNN on ADReSS and Canary train sets with prompt (\texttt{R+C+L}; \texttt{G.CoT}).}
    \label{fig:exp3}
    \vspace{-1.5ex}
\end{figure}

To evaluate the impact of $k$ in Delta-KNN, we systematically 
vary $k$ from $1$ to $20$ and perform cross-validation on the train sets of ADReSS and Canary. As shown in Figure~\ref{fig:exp3}, performance initially improves as $k$ increases, then declines and fluctuates. Empirically, we find that $k=13$ yields the best results on both datasets, so we adopt this value. 

We further examine the effect of $k$ under different prompts and observe varying optimal values. 
This suggests that determining the optimal number of target ``representatives'' in the Delta Matrix is non-trivial, as it possibly depends on multiple factors, including the prompt, language model, similarity computation, and text embedding model. As a result, 
determining the best $k$ requires a case-by-case approach.
For this reason, we rely on a held-out training set to empirically identify the best $k$. In future work, we aim to develop more advanced methods for optimizing this hyperparameter.

\section{Further Investigation}

We investigate various embedding methods (\cref{subsec:exp5}) and LLMs (\cref{subsec:exp6}), and finally compare our method against supervised classifiers (\cref{subsec:exp7}). 

\subsection{Delta-KNN using Other Text Encoders}
\label{subsec:exp5}

Beyond OpenAI embeddings, we investigate LLM hidden states as text representations, based on the assumption that the same LLM can better capture subtle linguistic nuances. 
We perform experiment with Llama-3.1-8B-Instruct using two common strategies: extracting embedding of an appended [EOS] token at the end of the text \citep{wang2023improving} and computing mean-pooled hidden states. 
Both approaches are applied at the first (L0), middle (L8, L16, L24), and the final layer (L32).  

\begin{figure}[t]
    \centering
    \includegraphics[width=.9\columnwidth]{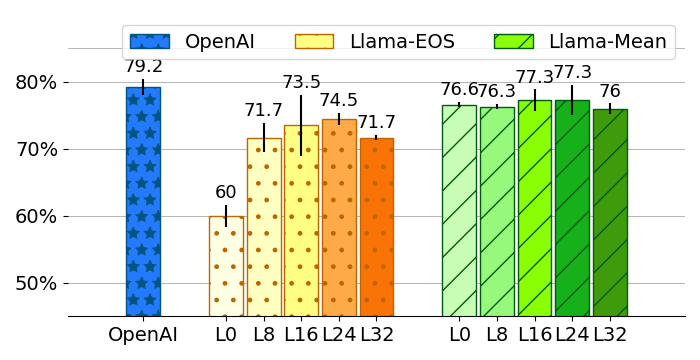}
    \caption{ADReSS-train performance using OpenAI \texttt{embedding-3-large} \textit{vs.} Llama-3.1-8B hidden states ([EOS] and mean-pooling) over different layers.
    }
    \label{fig:exp5}
    \vspace{-1.5ex}
\end{figure}

Figure~\ref{fig:exp5} presents the 4-shot ICL results on ADReSS-train using three encoding methods: OpenAI embeddings, [EOS] token and mean-pooled hidden states. Surprisingly, LLM-derived embeddings do not outperform external embeddings, with the best [EOS] and mean-pooled representations achieving $74.5\%$ and $77.3\%$ accuracy, respectively. 
Comparing the two approaches, we observe that mean-pooling provides more stable performance, while [EOS] embedding shows greater variance across different layers. 
The choice of layer also significantly impacts performance: 
mid-layers such as L16 and L24 outperform the last layer (L32), suggesting that mid layers encode richer semantic meaning, which is in line with \citet{chuang2024dola}. 
Presumably, a single layer's hidden states may capture only limited aspects of the input text. 
Future work could explore combining representations from multiple layers to enhance text encoding \citep{raymond2025detriever}.  
Additionally, we note recent advancements in transforming LLMs into effective text encoders, such as LLM2Vec \citep{behnamghader2024llm2vec}. Applying these methods could further boost the performance of Delta-KNN.

\subsection{Delta-KNN with Other LLMs}
\label{subsec:exp6}

We test the robustness of Delta-KNN on Mistral-7B-Instruct-v0.3
and Qwen2.5-7B-Instruct models.
The results in Table~\ref{tab:exp6} demonstrate that our method consistently outperforms other demonstration selection baselines across all tested LLMs, with Llama achieving the highest overall performance on both datasets.  
A closer analysis of performance variations across different prompts reveals that LLMs respond differently to the same instructions (detailed scores in Appendix~\ref{append:mistral-qwen}). 
In essence, Llama and Mistral perform best when provided with comprehensive prompts that include complete background information (\texttt{Role+Context+Linguistic}) and encourage step-by-step reasoning before making a prediction (\texttt{CoT}). 
In contrast, Qwen achieves its highest accuracy when prompted for a direct answer without explicit reasoning (\texttt{Role+Linguistic}). 
Interestingly, other demonstration selection methods also experience performance drops on Qwen when used more complex prompts, suggesting that prompt effectiveness is model-dependent. 
However, our approach remains robust and consistently improves performance across different prompting scenarios.

\begin{table}[t]
    \centering
    \resizebox{\columnwidth}{!}{
    \begin{tabular}{lccc}
    \toprule
     & ADReSS-train & ADReSS-test & Canary \\ 
    \midrule
    \rowcolor[gray]{0.9}\multicolumn{4}{l}{\textit{Mistral-7B-Instruct-v0.3}} \\
    Zero-shot & $52.3\textsubscript{0.5}$ & $67.7\textsubscript{1.0}$ & $63.1\textsubscript{0.8}$ \\
    Random & $62.0\textsubscript{2.8}$ & $70.8\textsubscript{2.1}$ & $55.0\textsubscript{0.4}$ \\
    Top-$k$ Select. & $53.2\textsubscript{2.3}$ & $63.5\textsubscript{3.1}$ & $62.3\textsubscript{0.0}$ \\
    ConE Select. & $61.1\textsubscript{1.9}$ & $66.7\textsubscript{4.2}$ & $58.8\textsubscript{3.5}$ \\
    Ours & $\textbf{69.9\textsubscript{1.4}}$ & $\textbf{76.0\textsubscript{5.2}}$ & $\textbf{72.3\textsubscript{0.4}}$ \\
    \midrule
    \rowcolor[gray]{0.9}\multicolumn{4}{l}{\textit{Qwen2.5-7B-Instruct}} \\
    Zero-shot & $61.6\textsubscript{0.5}$ & $66.8\textsubscript{2.2}$ & $63.5\textsubscript{0.4}$ \\
    Random & $62.0\textsubscript{2.8}$ & $57.3\textsubscript{1.0}$ & $64.6\textsubscript{3.8}$ \\
    Top-$k$ Select. & $58.8\textsubscript{1.4}$ & $66.7\textsubscript{2.1}$ & $53.1\textsubscript{6.2}$ \\
    ConE Select. & $58.8\textsubscript{0.5}$ & $65.8\textsubscript{5.3}$ & $60.0\textsubscript{1.5}$ \\
    Ours & $\textbf{63.4\textsubscript{0.5}}$ & $\textbf{67.7\textsubscript{0.0}}$ & $\textbf{66.1\textsubscript{2.7}}$ \\
    \bottomrule
    \end{tabular}}
    \caption{AD detection accuracy using Mistral and Qwen LLMs, with prompt (\texttt{Role+Context+Linguistic;} \texttt{G.-CoT}). The best score within each LLM is in \textbf{bold}.
    }
    \label{tab:exp6}
    \vspace{-2ex}
\end{table}

\subsection{Delta-KNN \textit{vs.} Supervised Baselines}
\label{subsec:exp7}

\begin{table}[]
    \centering
    \resizebox{\columnwidth}{!}{
    \begin{tabular}{lccc}
    \toprule
         & ADReSS-train & ADReSS-test & Canary\\
        \midrule
        \rowcolor[gray]{0.9}\multicolumn{4}{l}{\textit{Statistical ML Classifiers}} \\
        SVM \shortcite{luz2021detecting} & $80.7$ & $79.9$ & $51.9\textsubscript{3.5}$ \\
        NN \shortcite{balagopalan2021comparing} & $76.2$ & $77.1$ & - \\
        RF \shortcite{balagopalan2021comparing} & $73.8$ & $75.7$ & $68.7\textsubscript{1.9}$ \\
        LR \shortcite{jang2021classification} & - & - & $69.2\textsubscript{1.4}$ \\
        \midrule
        \rowcolor[gray]{0.9}\multicolumn{4}{l}{\textit{Transfer learning-based PLM}} \\
        BERT \shortcite{balagopalan2021comparing} & $\textbf{81.2\textsubscript{1.9}}$ $(^*81.8$) & $79.3\textsubscript{3.2}$ $(^*83.3$) & $71.7\textsubscript{2.6}$ \\
        GPT-3+SVM \shortcite{agbavor2022predicting} & $80.9$ &$80.3$ & -\\
        \midrule
        \rowcolor[gray]{0.9}\multicolumn{4}{l}{\textit{Fine-tuned LLM}} \\
        Llama-3.1-8B & $70.8\textsubscript{2.3}$ & $77.1\textsubscript{0.1}$ & $63.8\textsubscript{4.1}$ \\
        \midrule
        \rowcolor[gray]{0.9}\multicolumn{4}{l}{\textit{Learning-based ICL}} \\
        EPR \shortcite{rubin2022learning} & $79.5$ & $83.3$ & $66.9$ \\
        CEIL \shortcite{ye2023compositional} & $78.0$ & $82.1$ & $52.3$ \\
        \midrule
        \rowcolor[gray]{0.9}\multicolumn{4}{l}{\textit{Delta-KNN ICL}} \\
        Ours (Llama) & $80.0\textsubscript{1.3}$ & $\textbf{83.6\textsubscript{2.0}}$ & $\textbf{78.5\textsubscript{1.5}}$\\
        \bottomrule
    \end{tabular}}
    \caption{Accuracy comparison between supervised baselines and our method. Results are based on 10-fold cross-validation for ADReSS-train and Canary, and test set performance for ADReSS-test.
    On Canary, we re-implement SVM, RF, and LR following \citet{jang2021classification}. We fine-tune BERT (with $(^*)$ scores directly from \citet{balagopalan2021comparing}) and Llama. We re-implement learning-based retrieval methods EPR \citep{rubin2022learning} and CEIL \citep{ye2023compositional}.
    Best score per column is in \textbf{bold}. 
    }
    \label{tab:exp7}
    \vspace{-2ex}
\end{table}

Finally, we benchmark our Delta-KNN ICL with supervised baselines, see results in Table~\ref{tab:exp7}.

Traditional supervised methods, such as statistical machine learning and transfer-learning approaches, achieve good results. However, fine-tuning LLMs on this task does not lead to performance gains and instead underperforms compared to smaller supervised classifiers. 
This is expected, as our extremely small dataset likely lacks diversity and leads to overfitting \citep{garcia2023unreasonable}. This finding aligns with \citet{vieira2024much}, which shows that fine-tuning Llama on limited datasets ($1k$ samples) can degrade performance.  
For ICL, we benchmark with learning-based methods such as EPR \citep{rubin2022learning} and CEIL \citep{ye2023compositional}, where we fine-tune input and prompt encoders. 
Both methods perform well on ADReSS but lag behind on Canary, a more challenging cohort characterized by less and more diverse clinical data. We evaluate our Delta-KNN using model-optimized prompts, i.e., the best-performing prompt for each dataset, as shown in \cref{subsec:exp2}.
Excitingly, our approach achieves a new SOTA accuracy of $78.5\%$ on Canary while delivering competitive performance on ADReSS.
In addition, our search-based method is computationally more efficient, as it does not require fine-tuning any pre-trained or large language models.

Beyond strong performance, LLMs offers additional value by providing interpretable explanations that can assist doctors in diagnosis. To explore this, we conduct a qualitative study, where clinicians in our group compare LLM predictions with their own notes from a subset in Canary (Appendix~\ref{append:case}). Our findings suggest that LLMs strictly follow instructions and provide structured and insightful analyses, complementing human diagnosis.

\section{Conclusion}
\label{sec:conclusion}

We investigate the potential of LLMs as health assistants for AD detection, focusing on enhancing in-context learning. 
To tackle with limited data and the complexity of the task, we propose a novel demonstration selection method based on empirical evidence to quantify relative gains and identify optimal examples. 
Extensive experiments show that our approach consistently outperforms existing baselines, achieving substantial gains, 
particularly on the more challenging Canary dataset.

Moving forward, we intend to investigate alternative text encoding techniques and strategies for hyperparameter optimization. 
For probability score estimation, we prompt the LLM directly, but uncertainty-based methods like conditional entropy \citep{peng2024revisiting} offer interesting alternatives for future exploration.
Encouragingly, our method can be easily adapted for other data-poor scenarios and future applications such as integration with multimodal foundation models.

\section*{Limitations}

In terms of computation complexity:
(1) Constructing the Delta Matrix involves pairwise computations with a time complexity of $\mathcal{O}(n^2)$, where $n$ represents the number of training examples.
In practice, we leverage vLLM (\url{https://github.com/vllm-project/vllm}) for accelerated LLM inference. 
Given our small-data scenario, these computations remain feasible within standard computational resources.  
For prompts requiring only short answers (i.e., no CoT reasoning), inference for $10,000$ examples completes in approximately $10$ minutes. Prompts incorporating CoT reasoning take around $1.5$ hours. 
Notably, fine-tuning an LLM for just one epoch requires a similar runtime yet yields inferior results compared to our ICL approach.  
To scale our method to larger datasets, a possible solution is to apply clustering to the training examples, selecting a representative subset before constructing the Delta Matrix. 
(2) The second step, KNN selection, introduces only minimal computational overhead, with a linear complexity of $t \times \mathcal{O}(n)$, where $t$ denotes the number of test examples and $n$ the number of training examples. 
To enhance efficiency, we precompute and store the embeddings of all training examples, thereby eliminating the need for repeated computations.

For nearest neighbor selection (KNN), we explore multiple approaches, utilizing both an external text encoder from OpenAI and LLMs' internal hidden states. 
Our findings indicate that using LLM's inner embeddings does not enhance performance. 
However, improvements in similarity computation could be achieved through learning a similarity metric via contrastive learning or adopting advanced techniques to transform LLMs into more effective text encoders, such as LLM2Vec. 
Further advancements in this point could also help in optimizing the hyperparameter $k$.

Finally, we evaluate Delta-KNN across three LLMs from different families to assess the robustness of our approach. We focus on small-to-mid size models (7B–8B), balancing computational efficiency with strong performance. 
Testing our method on larger models or state-of-the-art open-source reasoning models, such as DeepSeek-R1, is an exciting direction for future exploration.

\section*{Ethical Considerations} 

The diagnosis of neurodegenerative disease is complex and relies on many indices. Automatic AI systems could provide clinicians with further clues, possibly alleviating the need for the patients to go through expensive and invasive screening tests, but this is a long-reach goal. 
In the healthcare domain, there is a risk that AI-generated predictions or analyses may be misinterpreted or directly relied upon as expert diagnoses. 
We emphasize the need for caution in their use. 
It is clear that the systems developed \textbf{can not} substitute for a human expert, as a diagnosis is a medical act. 
Moreover, linguistic cues and reasoning generated from LLMs, while helpful, have to be interpreted together with patient's clinical notes.

We carefully select the datasets used in this study to minimize potential biases and ensure that no private information---such as participants' health, clinical, or demographic data---is disclosed. 
This is a main reason for us exclusively testing with open-source language models. 
As authorized member of DementiaBank, we strictly follow its usage guidelines and ethical considerations. 
For the Canary dataset, 
the data collection process received approval from the Clinical Research Ethics Board at the University of British Columbia (H17-02803-A036).

The conception, implementation, analysis, and interpretation of results were conducted solely by the authors without any AI assistance. We used ChatGPT to check grammar.

\section*{Acknowledgments}
The authors thank the anonymous reviewers, Miini Teng for her input on the qualitative study, and the UBC Canary project team for their valuable feedback and suggestions.
The authors also extend their appreciation to all the trainees and research assistants who have contributed to the Canary project over the years by collecting and refining data.
This work was supported by the Faculty of Medicine and the Data Science Institute (DSI) at the University of British Columbia.
The authors acknowledge the support of the Natural Sciences and Engineering Research Council of Canada (NSERC).
Nous remercions le Conseil de recherches en sciences naturelles et en génie du Canada (CRSNG) de son soutien.
The computing resources are provided by the Digital Research Alliance of Canada (\url{https://www.alliancecan.ca/en}).

\bibliography{custom}

\appendix

\section{Data Statistics and Preprocessing}
\label{append:data-process}

\paragraph{Statistics.}
Table~\ref{tab:datasets2} shows the length (average and standard deviation of number of tokens, tokenized by BERT model), demographic (age, gender) and clinical (cognitive tests) information on ADReSS and Canary datasets.

Analyzing document length, we observe that healthy controls generally produce longer speech compared to AD patients, with more detailed descriptions and longer sentences.
However, documents in the Canary dataset are significantly longer than those in ADReSS and exhibit greater variation in length. The large variability suggests that Canary presents a more challenging dataset for AD detection.

\paragraph{Pre-processing.}
For ADReSS dataset, we extract clean texts by removing 
interviewer's content and special tokens such as non-verbal sounds encoded in the CHAT (Codes for Human Analysis of Transcripts) format.
We only use the textual transcripts.

For Canary dataset, 
participants completed four tasks--pupil calibration, picture description, paragraph reading, and memory recall--during which both language and eye movement data were collected. In this study, we only utilize data from the picture description task.
we use WhisperX \citep{bain2022whisperx} to first automatically transcribe the original speech data. The transcripts are manually verified by a human annotator to correct word spellings and speaker diarization.

\begin{table}[h!]
    \centering
    \resizebox{\columnwidth}{!}{
    \begin{tabular}{lcccc}
    \toprule
        & \multicolumn{2}{c}{ADReSS} & \multicolumn{2}{c}{Canary} \\
        \cmidrule(lr){2-3} \cmidrule(lr){4-5}
        & AD & Control & AD & Control \\
        \midrule
        \multicolumn{3}{l}{\textit{Training}}\\
        \hspace{0.7ex}\# Doc & $54$ & $54$ & $63$ & $67$ \\
        \hspace{0.7ex}Avg. \# Token & $122.2$ & $134.9$ & $150.9$ & $206.2$ \\
        \hspace{0.7ex}Std. \# Token & $76.2$ & $85.2$ & $102.5$ & $156.4$ \\
        \hspace{0.7ex}Age & $66.7\pm6.6$ & $66.4\pm6.5$ & $72\pm9$ & $62\pm15$ \\
        \hspace{0.7ex}Gender & $24$M / $30$F & $24$M / $30$F & $31$M / $34$F & $22$M / $45$F \\
        \hspace{0.7ex}Cognitive & $17.1\pm5.5$ & $29.1\pm1.0$ & $18\pm7$ & $27\pm3$ \\
        \midrule
        \multicolumn{3}{l}{\textit{Test}}\\
        \hspace{0.7ex}\# Doc & $24$ & $24$ & - & - \\
        \hspace{0.7ex}Avg. \# Token & $115.8$ & $154.9$ & - & - \\
        \hspace{0.7ex}Std. \# Token & $66.2$ & $107.6$ & - & - \\
        \hspace{0.7ex}Age & $66.1\pm7.4$ & $66.1\pm7.1$ & - & - \\
        \hspace{0.7ex}Gender & $13$M / $11$F & $13$M / $11$F & - & - \\
        \hspace{0.7ex}Cognitive & $19.5\pm5.4$ & $28.9\pm1.5$ & - & - \\
        \bottomrule
    \end{tabular}}
    \caption{Dataset demographic and clinical statistics. On cognitive tests, ADReSS reports Mini-Mental Status Examination score (MMSE); Canary reports Montreal Cognitive Assessment score (MoCA). - not applicable.}
    \label{tab:datasets2}
\end{table}

\section{Supervised Classifiers}
\label{append:superivsed}

Following \citet{jang2021classification}, we re-implement the supervised methods using Support Vector Machine (SVM), Logistic Regression (LR), and Random Forest (RF), all implemented with the Scikit-learn library \citep{pedregosa2011scikit}. 
To ensure robust evaluation, we perform $10$-fold cross-validation using ten different random seeds.
The average scores are given in Table~\ref{tab:supervised}, in comparison with our Delta-KNN results.

Note that noting that our results differ slightly from those reported in \citet{jang2021classification}, as we do not use the exact same training samples ($79$ Patients and $83$ Controls \textit{vs.} our dataset with $63$ Patients and $67$ Controls). Additionally, we employ different speech-to-text methods, which may have led to variations in the transcripts.

\begin{table}[] 
    \centering
    \resizebox{\columnwidth}{!}{
    \begin{tabular}{lcccc}
    \toprule
    Model & ACC & AUC & SEN & SPE \\
    \midrule
    SVM & $51.9\pm3.5$ & $43.3\pm6.8$ & $25.2\pm10.8$ & $79.7\pm5.4$ \\
    RF & $68.7\pm1.9$ & $73.6\pm1.9$ & $67.0\pm2.9$ & $70.2\pm4.0$ \\
    LR & $69.2\pm1.4$ & $73.6\pm1.4$ & $69.9\pm1.0$ & $68.3\pm2.3$ \\
    \midrule
    Ours & $78.5\pm1.5$ & $79.8\pm0.9$ & $70.6\pm0.8$ & $85.8\pm2.2$ \\
    \bottomrule
    \end{tabular}}
    \caption{Comparison of supervised classifiers (top) and our Delta-KNN ICL approach with Llama (bottom). RF: random forest, LR: logistic regression. Supervised results are averaged over $10$-seed $10$-fold cross-validation.
    }
    \label{tab:supervised}
\end{table}

\section{BERT Fine-tuning}
\label{append:ft-bert}

We fine-tune BERT on ADReSS-train and Canary following \citet{balagopalan2021comparing}, using the \texttt{bert-base-uncased} model \citep{devlin2018bert}.  
We use the [CLS] token from the final hidden state as the aggregate representation and pass it to the classification layer.  
BERT model is fine-tuned for $10$ epochs with a learning rate of $2e-5$, the same as in \citet{balagopalan2021comparing}.
Scores in Table~\ref{tab:exp7} regarding BERT are averaged over five runs.

\section{Llama Fine-tuning}
\label{append:ft-llama}

We also explore the feasibility of fine-tuning LLMs directly on our datasets. Given the extremely small size of our training data, we adopt Low-Rank Adaptation (LoRA) \citep{hulora2022}, a parameter-efficient fine-tuning approach. Specifically, we use low-rank (\(r=8\)) and low-alpha (\(\alpha=16\)) values while restricting updates to attention modules (\(Q, K, V, O\)) to mitigate overfitting.  

We fine-tune for a single epoch, as the training loss converges well, while additional epochs lead to a rebound in validation loss, indicating overfitting. All experiments are conducted on a single NVIDIA A100 40G GPU.  

For hyperparameter selection, we tested multiple configurations, including different rank values (8, 16), alpha values (16, 32), and target modules (``\textit{all-linear}'', ``\textit{q\_proj,v\_proj}'', ``\textit{q\_proj,k\_proj,v\_proj,o\_proj}''). Our results show that using lower alpha and dropout rates, combined with attention-only target modules, yields the best performance.
Detailed values for hyperparameters are presented in Table~\ref{tab:hyperparams}.

\begin{table}[]
    \centering
    \resizebox{\columnwidth}{!}{
    \begin{tabular}{lc}
    \toprule
    Hyperparameters & Selected \\
    \midrule
    \multicolumn{2}{l}{\textit{BitsAndBytes Quantisation}}\\
    \midrule
    use\_4bit\_quantization & True \\
    use\_nested\_quant & True \\
    bnb\_4bit\_compute\_dtype & bfloat16 \\
    \midrule
    \multicolumn{2}{l}{\textit{PEFT LoRA}}\\
    \midrule
    Lora r & $8$ \\
    Lora alpha & $16$ \\
    Lora dropout rate & $0.1$ \\
    Bias & none \\
    Task type & CAUSAL\_LM \\
    Target modules & q\_proj,k\_proj,v\_proj,o\_proj\\
    \midrule
    \multicolumn{2}{l}{\textit{Training Arguments}} \\
    \midrule
    Training epoch & $1$ \\
    Batch size & $1 $ \\
    Optimizer & adam \\
    Learning rate & $1e-4$ \\
    Learning rate scheduler & cosine\\
    Warm-up ratio & $0.0$ \\
    Weight decay & $1e-4$ \\
    \bottomrule
    \end{tabular}}
    \caption{Hyperparameters for Llama-3.1-8B-Instruct fine-tuning.}
    \label{tab:hyperparams}
\end{table}

\section{Prompt Templates}
\label{append:prompts}

We provide prompting template used in our experiments in Table~\ref{tab:temp-llama} on the next page.

\begin{table}[t]
    \centering
    \resizebox{\columnwidth}{!}{
    \begin{tabular}{llccc}
    \toprule
    & \multirow{3}{*}{Diagnosis} & \multirow{3}{*}{MoCA} & \multicolumn{2}{c}{Prediction} \\
    \cmidrule(lr){4-5}
    & & & LLM & Human \\ 
    \midrule
    Case 1 & AD & 3 & P & P \\
    Case 2 & Mild-moderate AD & 16 & P & P \\
    Case 3 & AD & 16 & P & P \\
    Case 4 & Mild AD & 25 & P & \textcolor{red}{H} \\
    Case 5 & aMCI$^*$ & 27 & P & P \\
    \midrule
    Case 6 & Healthy control & 21 & H & \textcolor{red}{P} \\
    Case 7 & Healthy control & 25 & \textcolor{red}{P} & H \\
    Case 8 & Healthy control & 28 & H & H \\
    Case 9 & Healthy control & 29 & H & H \\
    Case 10 & Healthy control & 30 & H & H \\
    \bottomrule
    \end{tabular}}
    \caption{Llama and human prediction on ten cases in Canary. aMCI$^*$: Amnestic mild cognitive impairment (aMCI). Prediction highlighted in red is incorrect.}
    \label{tab:qualitative}
\end{table}

\begin{table*}[]
    \centering
    \small
    \resizebox{\textwidth}{!}{
    \begin{tabular}{p{1.5cm} p{13cm}}
    \toprule
     & \small Template \\
    \midrule
    Background Prompt & \texttt{Role:} You are a medical expert in Alzheimer's disease. \\
    & \texttt{Context:} The Boston Cookie Theft picture description task is a well established speech assessment in Alzheimer's disease. During the task, participants are shown the picture and are asked to describe everything they see in the scene using as much time as they would like. The objects (also known as information units) in this picture includes: ``cookie'', ``girl'', ``boy'', ``woman'', ``jar'', ``stool'', ``plate'', ``dishcloth'', ``water'', ``window'', ``cupboard'', ``curtain'', ``dishes'', ``sink''. \vspace{1ex}\\
    & \texttt{Linguistic:} You analyze linguistic features in the patient's speech, such as lexical richness, syntactic complexity, grammatical correctness, information units, and semantic coherence. Based on the participant's description of the picture, provide an initial diagnosis of dementia patient (P) and healthy control (H).\\
    \midrule
    Example Prompt & \texttt{Zero-shot: None}\\
    & \texttt{Demonstration:} Example: \#\# Text: <text> \#\# Answer: healthy control (H). \#\# Text: <text> \#\# Answer: dementia patient (P). \\
    \midrule
    Question Prompt & 
    \texttt{CoT:} Given the text below, classify the participant as a dementia patient (P) or healthy control (H). First explain step-by-step and then give a prediction with a probability. \vspace{1ex} \\
    & \texttt{Guided CoT:} Given the text below, classify the participant as a dementia patient (P) or healthy control (H). Please first reason from the following perspectives: (1) Vocabulary richness: such as the usage of different words; (2) Syntactic complexity: such as the length of the sentence and the number of subordinate clauses; (3) Information content: whether the participant describe most of the information units in the picture; (4) Semantic coherence: such as the usage of connectives and the change in description from one information unit to another; (5) Fluency and repetitiveness: whether the text is fluent with less repetitive sentences. Based on your reasoning, please give a prediction and the corresponding probability.\\
    \bottomrule
    \end{tabular}}
    \caption{Prompt template used for AD detection.}
    \label{tab:temp-llama}
\end{table*}

\section{Results with Larger Number of In-Context Examples}
\label{append:80examples}

\begin{table*}[t]
    \centering
    \resizebox{\textwidth}{!}{
    \begin{tabular}{lcccccccccccc}
    \toprule
        \multirow{3}{*}{Method} & \multicolumn{4}{c}{ADReSS-train} & \multicolumn{4}{c}{ADReSS-test} &  \multicolumn{4}{c}{Canary}\\
        \cmidrule(lr){2-5} \cmidrule(lr){6-9} \cmidrule(lr){10-13}
        & 4-shot & 20-shot & 50-shot & 80-shot & 4-shot & 20-shot & 50-shot & 80-shot & 4-shot & 20-shot & 50-shot & 80-shot \\
        \midrule
    Random & $68.4\textsubscript{2.2}$ & $73.7\textsubscript{1.1}$ & $76.2\textsubscript{1.5}$ & $76.0\textsubscript{1.9}$ & $75.7\textsubscript{4.3}$ & $72.1\textsubscript{3.4}$ & $71.7\textsubscript{2.8}$ & $72.1\textsubscript{1.7}$ & $73.1\textsubscript{2.7}$ & $72.8\textsubscript{1.7}$ & $69.8\textsubscript{1.5}$ & $70.0\textsubscript{2.1}$ \\
    Top-$k$ Select. & $69.0\textsubscript{1.6}$ & $72.6\textsubscript{1.8}$ & $75.1\textsubscript{0.4}$ & $74.9\textsubscript{1.8}$ & $70.1\textsubscript{2.0}$ & $72.9\textsubscript{4.2}$ & $73.8\textsubscript{3.4}$ & $72.2\textsubscript{2.1}$ & $71.0\textsubscript{2.5}$ & $74.8\textsubscript{0.9}$ & $72.2\textsubscript{0.9}$ & $70.8\textsubscript{1.4}$ \\
    ConE$^*$ Select. & $67.4\textsubscript{2.3}$ & $75.7\textsubscript{1.1}$ & $75.8\textsubscript{1.2}$ & $75.1\textsubscript{1.7}$ & $70.1\textsubscript{1.0}$ & $75.9\textsubscript{2.1}$ & $73.0\textsubscript{2.6}$ & $71.7\textsubscript{2.8}$ & $73.3\textsubscript{1.9}$ & $72.2\textsubscript{1.0}$ & $70.9\textsubscript{1.0}$ & $70.6\textsubscript{1.2}$ \\
    \midrule
    Delta-KNN (ours) & $\mathbf{79.2\textsubscript{1.2}}$ & $76.1\textsubscript{1.1}$ & $78.4\textsubscript{2.4}$ & $76.6\textsubscript{0.9}$ & $\mathbf{80.5\textsubscript{3.9}}$ & $75.5\textsubscript{2.3}$ & $72.8\textsubscript{1.9}$ & $72.1\textsubscript{2.1}$ & $\mathbf{78.5\textsubscript{1.5}}$ & $74.7\textsubscript{1.1}$ & $72.2\textsubscript{1.2}$ & $70.8\textsubscript{1.0}$ \\
    \bottomrule
    \end{tabular}
    }
    \caption{
   Accuracy performance with 4-shot, 20-shot, 50-shot, and 80-shot in-context examples, comparing our method (using full prompt: \texttt{Role+Context+Linguistic}; \texttt{Demonstrations}; \texttt{G.-CoT}) against random sampling, Top-$k$ \citep{liu2022whatmakes}, and ConE$^*$-based \citep{peng2024revisiting} methods. 
    Best score per dataset is in \textbf{bold}.
    }
    \label{tab:80examples}
\end{table*}  

We experiment with extensive in-context examples ($N=20, 50, 80$), where $80$ examples are nearly the maximum number of demonstrations obtainable. 
The results with Llama-3.1-8B-Instruct model are given in Table~\ref{tab:80examples}, where the $4$-shot results are the same as in Table~\ref{tab:exp1}.
We see that extensive context does not guarantee superior results. 

Precisely, on the ADReSS-train dataset, where both the demonstration and test documents come from the same subset, we use 10-fold cross-validation. 
We observe some performance improvements with more examples on the baseline methods. However, they still fall short compared to our Delta-KNN 4-shot result.

On the ADReSS-test dataset, where the demonstration and test documents come from different subsets, increasing the number of shots only slightly improves the scores for Top-$k$ and ConE selection but results in lower performance with random sampling.

On the Canary dataset, which is the most diverse and challenging dataset, using a larger number of examples performs worse than the $4$-shot setting across all baselines.

These results suggest that while modern LLMs can process extremely long contexts, they struggle to focus on the most informative parts. 
Addressing this challenge is exactly the main objective of our paper: we aim to identify the most informative examples based on empirical evidence of relative gains (i.e., delta scores). 
Results show that our method can effectively select the examples to enhance LLMs performances on AD detection.

\section{Results with Mistral and Qwen}
\label{append:mistral-qwen}

We present AD prediction results using Mistral and Qwen with the prompt (\texttt{Role+Context+Linguistic}; \texttt{Demonstrations}; \texttt{G.-CoT}) in Table~\ref{tab:exp6mistral}, while results with other prompts are shown in Table~\ref{tab:exp6-1}, both on page 16.

As discussed in \cref{subsec:exp6}, different LLMs respond differently to the same prompt.  
Generally, more comprehensive prompts tend to yield better performance, as observed with Llama and Mistral. However, Qwen performs better with a simpler prompt. As shown in the last section of Table~\ref{tab:exp6-1}, Qwen achieves its highest accuracy when provided with \texttt{Role+Linguistic} and no CoT reasoning (prompt 3). When additional background information and CoT reasoning are introduced, its performance declines across all demonstration selection methods.

\section{Case Study on LLM's Prediction}
\label{append:case}

We conduct a qualitative study to examine how LLM-generated diagnoses compare with those made by a clinician in our research group. Specifically, we ask the clinician to provide diagnoses and reasoning for ten participants based solely on their picture description task outputs—without access to clinical notes—using similar instructions given to the LLM, see ``Instruction to human'' in Table~\ref{tab:case1} and Table~\ref{tab:case2}.  

Table~\ref{tab:qualitative} shows the predictions made by both the clinician and the LLM, alongside the ground-truth diagnoses and each participant’s Montreal Cognitive Assessment (MoCA) score. In this evaluation, the clinician correctly diagnosed eight cases, while Llama, utilizing Delta-KNN ICL, correctly identified nine. To illustrate the comparison in greater detail, we present the predictions and analyses from both the LLM and the clinician for Case 4 (Table~\ref{tab:case1}) and Case 7 (Table~\ref{tab:case2}).

In both cases, the clinician diagnosed the subjects as healthy controls, whereas Llama predicted them as patients. A closer analysis reveals that Llama follows a strictly structured approach by sequentially analyzing the input according to the Guided Chain-of-Thought prompt (G.-CoT, shown in Table~\ref{tab:temp-llama}) before summarizing its findings. In contrast, the clinician relies on pragmatic considerations, focusing on higher-level cognitive markers such as \textit{inference}, \textit{causality statements}, and \textit{logical event sequences}. However, this approach appears to overlook lower-level linguistic cues, such as lexical and syntactic patterns.  

The clinician’s diagnostic approach aligns more closely with human reasoning, as it highlights aspects that may be particularly revealing in assessing AD. Meanwhile, Llama’s analysis is systematic and precise, offering high readability and interpretability. Its diagnosis is directly rooted in the input text, providing detailed explanations for each aspect. For instance, it explicitly points out structural errors, such as: ``There are some errors in sentence structure, such as `And his mother is not really thinking about washing up because the water is running over the sink.'.''
This level of detailed reasoning and explanation could be valuable in assisting clinicians by offering an additional layer of linguistic analysis that might otherwise be overlooked.

\begin{table*}[t!]
    \centering
    \resizebox{\textwidth}{!}{
    \begin{tabular}{lllllllllllll}
    \toprule
        \multirow{3}{*}{Method} & \multicolumn{4}{c}{ADReSS-train} & \multicolumn{4}{c}{ADReSS-test} &  \multicolumn{4}{c}{Canary}\\
        \cmidrule(lr){2-5} \cmidrule(lr){6-9} \cmidrule(lr){10-13}
        & ACC & AUC & SEN & SPE & ACC & AUC & SEN & SPE & ACC & AUC & SEN & SPE \\
        \midrule
    \multicolumn{4}{l}{\textbf{Llama-3.1-8B-Instruct}}\\
    Zero-shot & $62.2\textsubscript{0.0}$ & $60.1\textsubscript{0.0}$ & $98.1\textsubscript{0.0}$ & $22.2\textsubscript{0.0}$ & $57.6\textsubscript{1.0}$ & $57.6\textsubscript{1.0}$ & $100.0\textsubscript{0.0}$ & $15.3\textsubscript{2.0}$ & $73.3\textsubscript{0.4}$ & $72.1\textsubscript{1.0}$ & $79.4\textsubscript{0.0}$ & $67.7\textsubscript{0.7}$ \\
    Random & $68.4\textsubscript{2.2}$ & $71.9\textsubscript{3.1}$ & $84.0\textsubscript{2.3}$ & $48.8\textsubscript{6.3}$ & $75.7\textsubscript{4.3}$ & $81.5\textsubscript{2.6}$ & $93.1\textsubscript{2.0}$ & $58.3\textsubscript{9.0}$ & $73.1\textsubscript{2.7}$ & $75.3\textsubscript{3.7}$ & $72.0\textsubscript{3.3}$ & $74.1\textsubscript{2.5}$ \\
    Top-$k$ Select. & $69.0\textsubscript{1.6}$ & $71.9\textsubscript{2.5}$ & $88.3\textsubscript{2.3}$ & $45.7\textsubscript{1.7}$ & $70.1\textsubscript{2.0}$ & $80.0\textsubscript{0.8}$ & $91.7\textsubscript{3.4}$ & $48.6\textsubscript{2.0}$ & $71.0\textsubscript{2.5}$ & $75.0\textsubscript{2.2}$ & $76.7\textsubscript{0.7}$ & $65.7\textsubscript{4.2}$ \\
    ConE$^*$ Select. & $67.4\textsubscript{2.3}$ & $74.5\textsubscript{1.3}$ & $85.2\textsubscript{1.5}$ & $45.7\textsubscript{3.1}$ & $70.1\textsubscript{1.0}$ & $76.4\textsubscript{2.6}$ & $93.1\textsubscript{2.0}$ & $47.2\textsubscript{2.0}$ & $73.3\textsubscript{1.9}$ & $78.4\textsubscript{0.9}$ & $79.9\textsubscript{2.0}$ & $67.2\textsubscript{4.4}$ \\
    Delta-KNN (ours) & $\underline{\mathbf{79.2\textsubscript{1.2}}}$ & $78.9\textsubscript{1.3}$ & $69.1\textsubscript{0.9}$ & $85.2\textsubscript{1.5}$ & $\underline{\textbf{80.5\textsubscript{3.9}}}$ & $85.8\textsubscript{0.9}$ & $70.8\textsubscript{5.9}$ & $86.1\textsubscript{2.0}$ & $\underline{\mathbf{78.5\textsubscript{1.5}}}$ & $79.8\textsubscript{0.9}$ & $70.6\textsubscript{0.8}$ & $85.8\textsubscript{2.2}$ \\
    \midrule
    
    \multicolumn{4}{l}{\textbf{Mistral-7B-Instruct-v0.3}}\\
    Zero-shot & $52.3\textsubscript{0.5}$ & $61.5\textsubscript{1.0}$ & $94.4\textsubscript{0.0}$ & $10.2\textsubscript{0.9}$ & $67.7\textsubscript{1.0}$ & $76.3\textsubscript{1.3}$ & $100.0\textsubscript{0.0}$ & $35.4\textsubscript{2.1}$ & $63.1\textsubscript{0.8}$ & $65.3\textsubscript{0.3}$ & $77.8\textsubscript{1.6}$ & $49.3\textsubscript{0.0}$ \\
    Random & $60.2\textsubscript{2.8}$ & $69.1\textsubscript{2.0}$ & $91.7\textsubscript{0.9}$ & $28.7\textsubscript{4.6}$ & $70.8\textsubscript{2.1}$ & $78.5\textsubscript{8.2}$ & $91.7\textsubscript{4.2}$ & $50.0\textsubscript{0.0}$ & $55.0\textsubscript{0.4}$ & $58.6\textsubscript{1.0}$ & $79.4\textsubscript{1.6}$ & $32.1\textsubscript{2.2}$ \\
    Top-$k$ & $53.2\textsubscript{2.3}$ & $69.6\textsubscript{4.2}$ & $88.9\textsubscript{5.6}$ & $17.6\textsubscript{0.9}$ & $63.5\textsubscript{3.1}$ & $74.0\textsubscript{1.6}$ & $91.7\textsubscript{0.0}$ & $35.4\textsubscript{6.2}$ & $62.3\textsubscript{0.0}$ & $69.2\textsubscript{1.9}$ & $84.9\textsubscript{2.4}$ & $41.0\textsubscript{2.2}$ \\
    ConE$^*$ Select. & $61.1\textsubscript{1.9}$ & $78.0\textsubscript{3.9}$ & $93.5\textsubscript{2.8}$ & $28.7\textsubscript{0.9}$ & $66.7\textsubscript{4.2}$ & $78.0\textsubscript{0.7}$ & $97.9\textsubscript{2.1}$ & $35.4\textsubscript{6.2}$ & $58.8\textsubscript{3.5}$ & $64.2\textsubscript{2.3}$ & $84.9\textsubscript{2.4}$ & $34.3\textsubscript{4.5}$ \\
    Delta-KNN (ours) & $\underline{69.9\textsubscript{1.4}}$ & $82.4\textsubscript{3.2}$ & $90.7\textsubscript{0.0}$ & $49.1\textsubscript{2.8}$ & $\underline{76.0\textsubscript{5.2}}$ & $84.9\textsubscript{2.7}$ & $95.8\textsubscript{0.0}$ & $56.2\textsubscript{10.4}$ & $\underline{72.3\textsubscript{0.4}}$ & $74.8\textsubscript{0.2}$ & $86.5\textsubscript{0.8}$ & $59.0\textsubscript{0.7}$ \\
    \midrule
    
    \multicolumn{4}{l}{\textbf{Qwen2.5-7B-Instruct}}\\
    Zero-shot & $61.6\textsubscript{0.5}$ & $64.9\textsubscript{2.9}$ & $94.4\textsubscript{0.0}$ & $28.7\textsubscript{0.9}$ & $66.8\textsubscript{2.2}$ & $65.5\textsubscript{2.6}$ & $97.9\textsubscript{2.1}$ & $43.8\textsubscript{10.4}$ &  
    $63.5\textsubscript{0.4}$ & $62.5\textsubscript{0.5}$ & $69.0\textsubscript{0.8}$ & $58.2\textsubscript{1.5}$ \\
    Random & $62.0\textsubscript{2.8}$ & $62.5\textsubscript{2.2}$ & $89.8\textsubscript{2.8}$ & $34.3\textsubscript{2.8}$ & $57.3\textsubscript{1.0}$ & $53.8\textsubscript{0.7}$ & $75.0\textsubscript{4.2}$ & $39.6\textsubscript{2.1}$ & 
    $64.6\textsubscript{3.8}$ & $63.2\textsubscript{3.8}$ & $81.0\textsubscript{4.8}$ & $49.3\textsubscript{3.0}$ \\
    Top-$k$ Select. & $58.8\textsubscript{1.4}$ & $56.2\textsubscript{0.0}$ & $88.0\textsubscript{2.8}$ & $29.6\textsubscript{0.0}$ & $66.7\textsubscript{2.1}$ & $65.5\textsubscript{6.0}$ & $91.7\textsubscript{0.0}$ & $41.7\textsubscript{4.2}$ & 
    $53.1\textsubscript{6.2}$ & $51.6\textsubscript{7.1}$ & $70.6\textsubscript{5.6}$ & $36.6\textsubscript{6.7}$ \\
    ConE$^*$ Select. & $58.8\textsubscript{0.5}$ & $58.9\textsubscript{1.4}$ & $88.0\textsubscript{2.8}$ & $29.6\textsubscript{1.9}$ & $65.8\textsubscript{5.3}$ & $63.8\textsubscript{12.7}$ & $91.7\textsubscript{8.3}$ & $45.8\textsubscript{8.3}$ & 
    $60.0\textsubscript{1.5}$ & $57.4\textsubscript{0.0}$ & $68.3\textsubscript{7.9}$ & $52.2\textsubscript{4.5}$ \\
    Delta-KNN (ours) & $\underline{63.4\textsubscript{0.5}}$ & $62.7\textsubscript{2.2}$ & $82.4\textsubscript{0.9}$ & $44.4\textsubscript{0.0}$ & $\underline{67.7\textsubscript{0.0}}$ & $62.2\textsubscript{1.1}$ & $85.4\textsubscript{6.2}$ & $47.9\textsubscript{6.2}$ &
    $\underline{66.1\textsubscript{2.7}}$ & $64.8\textsubscript{3.9}$ & $71.4\textsubscript{0.0}$ & $45.5\textsubscript{5.2}$ \\
    \bottomrule
    \end{tabular}
    }
    \caption{AD detection results using different demonstration selection methods on Llama, Mistral, and Qwen models; prompt (\texttt{Role+Context+Linguistic;} \texttt{G.-CoT}). 
    The best accuracy within each LLM is \underline{underlined} while the overall highest accuracy is in \textbf{bold}.
    }
    \label{tab:exp6mistral}
\end{table*}

\begin{table*}[t!]
    \centering
    \resizebox{\textwidth}{!}{
    \begin{tabular}{ccccccllllll}
        \toprule
        & \multirow{3}{*}{Role} & \multirow{3}{*}{Con.} & \multirow{3}{*}{Ling.} & \multirow{3}{*}{CoT} & \multirow{3}{*}{G.-CoT} & \multicolumn{2}{c}{ADReSS-train} & \multicolumn{2}{c}{ADReSS-test} & \multicolumn{2}{c}{Canary} \\
        \cmidrule(lr){7-12}
        && & & & & \multicolumn{2}{c}{Delta-KNN (Rdm, Top$k$, ConE)} & \multicolumn{2}{c}{Delta-KNN (Rdm, Top$k$, ConE)} & \multicolumn{2}{c}{Delta-KNN (Rdm, Top$k$, ConE)} \\
        \midrule
        \multicolumn{4}{l}{\textbf{Llama-3.1-8B-Instruct}}\\
        (1) & \textcolor{red}{\ding{55}} & \textcolor{red}{\ding{55}} & \textcolor{red}{\ding{55}} & \textcolor{red}{\ding{55}} & \textcolor{red}{\ding{55}} 
        & $73.0$ &
        $\downarrow13.9$  
        $\downarrow17.6$
        $\downarrow12.7$
        & $69.8$ & $\uparrow0.3$ \hspace{2.72pt} $\downarrow2.8$ \hspace{2.72pt} $\uparrow0.8$
        & $63.1$ & $\downarrow2.3$ \hspace{2.72pt} $\downarrow2.3$ \hspace{2.72pt} $\downarrow3.9$ \\
        (2) & 
        \textcolor{blue}{\ding{51}} & \textcolor{blue}{\ding{51}} & \textcolor{red}{\ding{55}} & \textcolor{red}{\ding{55}} & \textcolor{red}{\ding{55}} 
        & $72.7$ & $\downarrow2.1$ \hspace{2.72pt} $\downarrow2.4$ \hspace{2.72pt} $\downarrow3.3$
        & $69.1$ & $\sim 0$ \hspace{8.72pt} $\sim 0$ \hspace{7.75pt} $\downarrow2.9$ 
        & $70.0$ & $\downarrow3.8$ \hspace{2.72pt} $\downarrow3.8$ \hspace{2.72pt} $\downarrow4.1$ \\
        (3) & 
        \textcolor{blue}{\ding{51}} & \textcolor{red}{\ding{55}} & \textcolor{blue}{\ding{51}} & \textcolor{red}{\ding{55}} & \textcolor{red}{\ding{55}} 
        & $73.1$ & $\downarrow7.8$ \hspace{2.72pt} $\downarrow13.1$  $\downarrow6.9$ 
        & $74.4$ & $\downarrow5.3$ \hspace{2.72pt} $\downarrow2.6$ \hspace{2.72pt} $\downarrow1.2$
        & $68.1$ & $\downarrow9.1$ \hspace{2.72pt} $\downarrow2.2$ \hspace{2.72pt} $\downarrow2.7$ \\
        (4) & 
        \textcolor{blue}{\ding{51}} & \textcolor{blue}{\ding{51}} & \textcolor{red}{\ding{55}} & \textcolor{blue}{\ding{51}} & \textcolor{red}{\ding{55}} 
        & $73.6$ & $\downarrow5.9$ \hspace{2.72pt} $\downarrow4.9$ \hspace{2.72pt} $\downarrow6.3$  
        & $74.6$ & $\downarrow2.1$ \hspace{2.72pt} $\downarrow2.8$ \hspace{2.72pt} $\downarrow2.8$ 
        & $71.5$ & $\downarrow9.7$ \hspace{2.72pt} $\downarrow4.3$ \hspace{2.72pt} $\downarrow10.2$ \\
        (5) & 
        \textcolor{blue}{\ding{51}} & \textcolor{red}{\ding{55}} & \textcolor{blue}{\ding{51}} & \textcolor{blue}{\ding{51}} & \textcolor{red}{\ding{55}} 
        & $74.5$ & $\downarrow10.2$ $\downarrow14.5$ $\downarrow16.3$
        & $74.6$ & $\downarrow11.1$  $\downarrow13.2$ $\downarrow16.7$ 
        & $65.1$ & $\downarrow3.6$ \hspace{2.72pt} $\downarrow6.9$ \hspace{2.72pt} $\downarrow7.7$ \\
        (6) & 
        \textcolor{blue}{\ding{51}} & \textcolor{blue}{\ding{51}} & \textcolor{blue}{\ding{51}} & \textcolor{blue}{\ding{51}} & \textcolor{red}{\ding{55}} 
        & $80.0$ & $\downarrow9.9$ \hspace{2.72pt} $\downarrow11.1$ $\downarrow8.6$ 
        & $83.6$ & $\downarrow13.8$  $\downarrow10.4$ $\downarrow12.5$ 
        & $70.8$ & $\downarrow7.5$ \hspace{2.72pt} $\downarrow7.7$ \hspace{2.72pt} $\downarrow9.6$ \\
        (7) & 
        \textcolor{blue}{\ding{51}} & \textcolor{blue}{\ding{51}} & \textcolor{blue}{\ding{51}} & \textcolor{red}{\ding{55}} & \textcolor{blue}{\ding{51}}
        & $79.2$ & $\downarrow10.8$ $\downarrow10.2$ $\downarrow11.8$ 
        & $80.5$ & $\downarrow2.8$ \hspace{2.72pt} $\downarrow8.4$ \hspace{2.72pt} $\downarrow8.4$ 
        & $78.5$ & $\downarrow5.4$ \hspace{2.72pt} $\downarrow7.5$ \hspace{2.72pt} $\downarrow5.2$ \\
        \midrule
        
        \multicolumn{4}{l}{\textbf{Mistral-7B-Instruct-v0.3}}\\
        (1) & \textcolor{red}{\ding{55}} & \textcolor{red}{\ding{55}} & \textcolor{red}{\ding{55}} & \textcolor{red}{\ding{55}} & \textcolor{red}{\ding{55}} 
        & $50.0$ & $\sim0$ \hspace{8.72pt} $\sim0$ \hspace{7.75pt} $\sim0$ & $51.0$ & $\downarrow1.0$ \hspace{2.72pt} $\downarrow1.0$ \hspace{2.72pt} $\sim0$ & $51.2$ & $\downarrow2.7$ \hspace{2.72pt} $\downarrow1.5$ \hspace{2.72pt} $\downarrow1.2$ \\
        (2) & \textcolor{blue}{\ding{51}} & \textcolor{blue}{\ding{51}} & \textcolor{red}{\ding{55}} & \textcolor{red}{\ding{55}} & \textcolor{red}{\ding{55}} 
        & $51.4$ & $\downarrow1.4$ \hspace{2.72pt} $\downarrow1.4$ \hspace{2.72pt} $\downarrow1.4$ & $53.1$ & $\downarrow0.9$ \hspace{2.72pt} $\downarrow2.1$ \hspace{2.72pt} $\downarrow3.1$ & $55.0$ & $\downarrow6.5$ \hspace{2.72pt} $\downarrow6.5$ \hspace{2.72pt} $\downarrow6.5$ \\
        (3) & \textcolor{blue}{\ding{51}} & \textcolor{red}{\ding{55}} & \textcolor{blue}{\ding{51}} & \textcolor{red}{\ding{55}} & \textcolor{red}{\ding{55}} 
        & $50.5$ & $\uparrow0.4$ \hspace{2.72pt} $\downarrow0.5$ \hspace{2.72pt} $\downarrow0.5$ & $55.2$ & $\downarrow4.2$ \hspace{2.72pt} $\downarrow5.2$ \hspace{2.72pt} $\downarrow5.2$ & $49.6$ & $\downarrow0.4$ \hspace{2.72pt} $\downarrow0.9$ \hspace{2.72pt} $\downarrow0.9$ \\
        (4) & \textcolor{blue}{\ding{51}} & \textcolor{blue}{\ding{51}} & \textcolor{red}{\ding{55}} & \textcolor{blue}{\ding{51}} & \textcolor{red}{\ding{55}} 
        & $63.0$ & $\downarrow4.2$ \hspace{2.72pt} $\downarrow2.8$ \hspace{2.72pt} $\downarrow4.7$ & $68.8$ & $\downarrow7.3$ \hspace{2.72pt} $\downarrow4.3$ \hspace{2.72pt} $\downarrow10.5$ & $63.1$ & $\downarrow7.2$ \hspace{2.72pt} $\downarrow13.5$ $\downarrow12.7$ \\
        (5) & \textcolor{blue}{\ding{51}} & \textcolor{red}{\ding{55}} & \textcolor{blue}{\ding{51}} & \textcolor{blue}{\ding{51}} & \textcolor{red}{\ding{55}} 
        & $58.8$ & $\sim0$ \hspace{7.75pt} $\downarrow9.7$ \hspace{2.72pt} $\downarrow1.4$ & $65.6$ & $\downarrow4.1$ \hspace{2.72pt} $\downarrow10.4$ \hspace{0.5pt} $\sim0$ & $65.4$ & $\downarrow10.2$ $\downarrow9.8$ \hspace{2.72pt} $\downarrow6.2$\\
        (6) & \textcolor{blue}{\ding{51}} & \textcolor{blue}{\ding{51}} & \textcolor{blue}{\ding{51}} & \textcolor{blue}{\ding{51}} & \textcolor{red}{\ding{55}} 
        & $68.5$ & $\downarrow9.2$ \hspace{2.72pt} $\downarrow11.1$ $\downarrow6.9$ & $79.2$ & $\downarrow15.7$ $\downarrow15.7$ $\downarrow19.8$ & $65.0$ & $\downarrow4.5$ \hspace{2.72pt} $\downarrow7.7$ \hspace{2.72pt} $\downarrow7.7$\\
        (7) & \textcolor{blue}{\ding{51}} & \textcolor{blue}{\ding{51}} & \textcolor{blue}{\ding{51}} & \textcolor{red}{\ding{55}} & \textcolor{blue}{\ding{51}}
        & $69.9$ & $\downarrow9.7$ \hspace{2.72pt} $\downarrow16.7$ $\downarrow8.7$ & $76.0$ & $\downarrow5.2$ \hspace{2.72pt} $\downarrow12.5$ $\downarrow9.3$ & $72.3$ & $\downarrow17.3$ $\downarrow10.0$ $\downarrow13.5$\\
        \midrule

        \multicolumn{4}{l}{\textbf{Qwen2.5-7B-Instruct}}\\
        (1) & \textcolor{red}{\ding{55}} & \textcolor{red}{\ding{55}} & \textcolor{red}{\ding{55}} & \textcolor{red}{\ding{55}} & \textcolor{red}{\ding{55}} 
        & $66.7$ & $\downarrow6.7$ \hspace{2.72pt} $\downarrow11.6$ $\downarrow7.4$ & $72.9$ & $\downarrow1.0$ \hspace{2.72pt} $\downarrow8.3$ \hspace{2.72pt} $\downarrow12.5$ & $67.7$ & $\downarrow6.2$ \hspace{2.72pt} $\downarrow10.8$ $\downarrow4.2$\\
        (2) & \textcolor{blue}{\ding{51}} & \textcolor{blue}{\ding{51}} & \textcolor{red}{\ding{55}} & \textcolor{red}{\ding{55}} & \textcolor{red}{\ding{55}} 
        & $70.8$ & $\downarrow7.2$ \hspace{2.72pt} $\downarrow11.5$ $\downarrow6.0$ & $71.9$ & $\downarrow6.3$ \hspace{2.72pt} $\sim0$ \hspace{7.72pt} $\downarrow6.3$ & $67.3$ & $\downarrow4.2$ \hspace{2.72pt} $\downarrow3.1$ \hspace{2.72pt} $\downarrow0.5$ \\
        (3) & \textcolor{blue}{\ding{51}} & \textcolor{red}{\ding{55}} & \textcolor{blue}{\ding{51}} & \textcolor{red}{\ding{55}} & \textcolor{red}{\ding{55}} 
        & $69.4$ & $\downarrow4.6$ \hspace{2.72pt} $\downarrow6.6$ \hspace{2.72pt} $\downarrow7.4$ & $78.1$ & $\downarrow3.1$ \hspace{2.72pt} $\downarrow5.3$ \hspace{2.72pt} $\downarrow10.4$ & $69.2$ & $\downarrow3.6$ \hspace{2.72pt} $\downarrow6.5$ \hspace{2.72pt} $\downarrow4.2$\\
        (4) & \textcolor{blue}{\ding{51}} & \textcolor{blue}{\ding{51}} & \textcolor{red}{\ding{55}} & \textcolor{blue}{\ding{51}} & \textcolor{red}{\ding{55}} 
        & $70.0$ & $\uparrow1.8$ \hspace{2.72pt} $\downarrow4.7$ \hspace{2.72pt} $\downarrow1.9$ & $69.7$ & $\downarrow7.2$ \hspace{2.72pt} $\uparrow1.2$ \hspace{2.72pt} $\downarrow4.2$ & $61.9$ & $\uparrow0.4$ \hspace{2.72pt} $\downarrow1.5$ \hspace{2.72pt} $\uparrow0.6$ \\
        (5) & \textcolor{blue}{\ding{51}} & \textcolor{red}{\ding{55}} & \textcolor{blue}{\ding{51}} & \textcolor{blue}{\ding{51}} & \textcolor{red}{\ding{55}} 
        & $63.9$ & $\downarrow4.2$ \hspace{2.72pt} $\downarrow0.5$ \hspace{2.72pt} $\downarrow8.3$ & $76.0$ & $\downarrow7.2$ \hspace{2.72pt} $\downarrow6.8$ \hspace{2.72pt} $\downarrow6.8$ & $63.1$ & $\uparrow5.4$ \hspace{2.72pt} $\downarrow4.6$ \hspace{2.72pt} $\downarrow2.2$ \\
        (6) & \textcolor{blue}{\ding{51}} & \textcolor{blue}{\ding{51}} & \textcolor{blue}{\ding{51}} & \textcolor{blue}{\ding{51}} & \textcolor{red}{\ding{55}} 
        & $59.7$ & $\downarrow2.6$ \hspace{2.72pt} $\uparrow2.8$ \hspace{2.72pt} $\downarrow0.9$ & $64.6$ & $\downarrow3.1$ \hspace{2.72pt} $\downarrow2.1$ \hspace{2.72pt} $\downarrow0.9$ & $56.9$ & $\downarrow8.7$ \hspace{2.72pt} $\downarrow4.0$ \hspace{2.72pt} $\downarrow4.5$ \\
        (7) & \textcolor{blue}{\ding{51}} & \textcolor{blue}{\ding{51}} & \textcolor{blue}{\ding{51}} & \textcolor{red}{\ding{55}} & \textcolor{blue}{\ding{51}}
        & $63.4$ & $\downarrow2.4$ \hspace{2.72pt} $\downarrow4.6$ \hspace{2.72pt} $\downarrow4.6$ & $67.7$ & $\downarrow10.4$ $\downarrow1.0$ \hspace{2.72pt} $\downarrow1.9$ & $66.1$ & $\downarrow1.5$ \hspace{2.72pt} $\downarrow13.0$ $\downarrow6.1$\\
        \midrule
        \bottomrule
    \end{tabular}
    }
    \caption{
    Delta-KNN performance (accuracy) using different prompt engineering strategies (Role, Context, Linguistic cues, chain-of-thought reasoning (CoT), and Guided CoT) on ADReSS and Canary datasets, in comparison with Random sampling (Rdm), Top-$k$~\citep{liu2022whatmakes} selection, and Conditional Entropy (ConE)~\citep{peng2024revisiting} selection. 
    $\downarrow$, $\uparrow$, and $\sim$ symbols refer to lower, higher, and same accuracies compared to the Delta-KNN, respectively.
    }
    \label{tab:exp6-1}
\end{table*}

\begin{table*}[]
    \centering
    \small
    \resizebox{\textwidth}{!}{
    \begin{tabular}{p{1.5cm} p{13cm}}
    \toprule
    & \small Case study 4 (\textcolor{blue}{Gold label: P})\\
    \midrule
    Instruction to Human &
    This experiment aims to explore how clinicians assess language in the Cookie Theft task and identify the key criteria (reasoning processes) they use to make judgments.  
    Below is a transcript produced by a participant (either a dementia patient or a healthy control). Based on the participant's description of the picture, please provide a brief analysis (2-10 sentences). Then, make a prediction by classifying the participant as either a healthy control (H) or a dementia patient (P). \\
    \midrule
    Instruction to LLM &  See prompt template in Table~\ref{tab:temp-llama}. \\
    \midrule
    \midrule
    Input Text & ``There's a boy that's stealing cookies from the cookie jar. And his mother is not really thinking about washing up because the water is running over the sink. And the boy is going to fall because of the stool. And the only thing his sister is interested in is for him to get more cookies, throw more cookies around. And this looks like a driveway outside. And there's some cups and two cups and a plate. And I don't know, anything of note? No?''\\
    \midrule
    Human & \textcolor{blue}{Healthy control}. \\
    Diagnosis & There were \textbf{inferences} made from the picture, such as identifying individuals and assuming their relationships (e.g., recognizing a mother, a boy, and a girl). Moreover, the participant was making \textbf{causality statements}, such as predicting that the boy is going to fall because of the unstable stool. \\
    \midrule
    Llama Diagnosis & Here's the step-by-step analysis: \textbf{1. Lexical richness}: The participant uses a variety of words, including ``stealing'', ``interested'', and ``driveway'', which suggests a good vocabulary. \textbf{2. Syntactic complexity}: The participant's sentences are relatively simple, with a mix of short and longer sentences. However, there are some errors in sentence structure, such as ``And his mother is not really thinking about washing up because the water is running over the sink.'' This sentence is a bit convoluted and could be rephrased more clearly. \textbf{3. Grammatical correctness}: The participant makes a few grammatical errors, such as using ``And'' instead of a more formal conjunction, and using a comma splice in the sentence ``And the only thing his sister is interested in is for him to get more cookies, throw more cookies around.'' \textbf{4. Information units}: The participant mentions most of the objects in the picture, including the boy, mother, sister, cookie jar, stool, sink, cups, and plate. However, they seem to be having trouble with the window and the scene outside, which they incorrectly identify as a driveway. \textbf{5. Semantic coherence}: The participant's description is generally coherent, but there are some strange inferences, such as the sister being interested in the boy throwing cookies around. Based on these observations, I would classify the participant as a \textcolor{blue}{dementia patient (P)} with a probability of \textcolor{blue}{0.7}. The participant's description is generally clear and coherent, but there are some errors in sentence structure, grammar, and semantic coherence that suggest cognitive impairment. However, the participant's vocabulary and ability to identify most of the objects in the picture suggest that they are not severely impaired. \\
    \bottomrule
    \end{tabular}}
    \caption{The first example of a comparison between Llama's and human's prediction on participant 4's speech. Human's prediction is incorrect, while Llama's prediction is correct.}
    \label{tab:case1}
\end{table*}

\begin{table*}[]
    \centering
    \small
    \resizebox{\textwidth}{!}{
    \begin{tabular}{p{1.5cm} p{13cm}}
    \toprule
    & \small Case study 7 (\textcolor{blue}{Gold label: H})\\
    \midrule
    Instruction to Human &
    This experiment aims to explore how clinicians assess language in the Cookie Theft task and identify the key criteria (reasoning processes) they use to make judgments.  
    Below is a transcript produced by a participant (either a dementia patient or a healthy control). Based on the participant's description of the picture, please provide a brief analysis (2-10 sentences). Then, make a prediction by classifying the participant as either a healthy control (H) or a dementia patient (P). \\
    \midrule
    Instruction to LLM &  See prompt template in Table~\ref{tab:temp-llama}. \\
    \midrule
    \midrule
    Input Text & ``There is a boy trying to reach a cookie jar, falling off a stool almost. His sister reaching up for a cookie. His mother is washing the dishes. And the sink is overflowing with water. And I'm finished.''\\
    \midrule
    Human & \textcolor{blue}{Healthy control.} \\
    Diagnosis & This person was making lots of role assumptions, with \textbf{logical sequence} of events (from dishes to water overflowing). \\
    \midrule
    Llama Diagnosis & Here's the step-by-step analysis: \textbf{1. Lexical richness}: The participant uses a limited vocabulary, with only 7 words or phrases to describe the scene. This is a characteristic of dementia patients, who often experience word-finding difficulties. \textbf{2. Syntactic complexity}: The participant's sentence structure is simple and lacks complexity. They use short, basic sentences with no subordinate clauses or embedded phrases. This is consistent with the language patterns of dementia patients. \textbf{3. Grammatical correctness}: The participant's sentences are grammatically correct, but there is a lack of detail and elaboration. This is a characteristic of dementia patients, who may struggle to provide nuanced descriptions. \textbf{4. Information units}: The participant identifies 5 key elements in the scene: the boy, the sister, the mother, the cookie jar, and the overflowing sink. However, they fail to mention other important details, such as the stool, the sister's actions, and the mother's apron. \textbf{5. Semantic coherence}: The participant's description is disjointed and lacks coherence. They jump abruptly from describing the boy's actions to stating that they are finished. This is a characteristic of dementia patients, who may struggle to maintain a coherent narrative. Based on these observations, I predict that the participant is a \textcolor{blue}{dementia patient (P)} with a high probability of \textcolor{blue}{0.85}. The participant's language patterns, including their limited vocabulary, simple sentence structure, and lack of detail, are consistent with the characteristics of dementia patients. \\
    \bottomrule
    \end{tabular}}
    \caption{The second example of a comparison between Llama's and human's prediction on participant 7's speech. Human's prediction is correct, while Llama's prediction is incorrect.}
    \label{tab:case2}
\end{table*}

\end{document}